\documentclass[12pt]{article}

\usepackage{newtxtext,newtxmath}

\usepackage{graphicx}

\usepackage[letterpaper,margin=1in]{geometry}

\renewenvironment{abstract}
	{\quotation}
	{\endquotation}

\date{}

\makeatletter
\renewcommand{\fnum@figure}{\textbf{Figure \thefigure}}
\renewcommand{\fnum@table}{\textbf{Table \thetable}}
\makeatother

\usepackage{scicite}

\usepackage{url}

\def\eg{{\em {\em e.g.},\ }}
\def\ie{{\em {\em i.e.},\ }}

\def\crawlgait{\emph{Crawl~N'~Sense}}
\def\trotgait{\emph{Trot-walk}}

\def\scititle{
	Legged Autonomous Surface Science In Analogue Environments (LASSIE): Making Every Robotic Step Count in Planetary Exploration
}
\title{\bfseries \boldmath \scititle}

\author{
Cristina G. Wilson$^{1\dagger}$, Marion Nachon$^{2\dagger}$, Shipeng Liu$^{3}$, John G. Ruck$^{4}$,  
\and J. Diego Caporale$^{3}$, Benjamin E. McKeeby$^{5}$, Yifeng Zhang$^{3}$, Jordan M. Bretzfelder$^{6}$, 
\and John Bush$^{3}$, Alivia M. Eng$^{6}$, Ethan Fulcher$^{3}$, Emmy B. Hughes$^{6}$, Ian C. Rankin$^{1}$, 
\and Jelis J. Sostre Cortés$^{6}$, Sophie Silver$^{4}$, Michael R. Zanetti$^{7}$, Ryan C. Ewing$^{8}$,
\and Kenton R. Fisher$^{8}$, Douglas J. Jerolmack$^{4}$, Daniel E. Koditschek$^{9}$,
\and Frances Rivera-Hernández$^{6}$, Thomas F. Shipley$^{10}$, and Feifei Qian$^{3\ast}$
\and
\small{$^{1}$Collaborative Robotics and Intelligent Systems Institute, Oregon State University, Corvallis, 97331, USA}\and
\small{$^{2}$Department of Geology and Geophysics, Texas A\&M University, College Station, 77843, USA}\and
\small{$^{3}$Department of Electrical and Computer Engineering, University of Southern California, Los Angeles, 90089, USA}\and
\small{$^{4}$Department of Earth and Environmental Science, University of Pennsylvania, Philadelphia, 19104, USA}\and
\small{$^{5}$Department of Astronomy and Planetary Science, Northern Arizona University, Flagstaff, 86011, USA}\and
\small{$^{6}$School of Earth and Atmospheric Sciences, Georgia Institute of Technology, Atlanta, 30332, USA}\and
\small{$^{7}$NASA Marshall Space Flight Center, Huntsville, 35812, USA}\and
\small{$^{8}$NASA Johnson Space Center, Houston, 77058, USA}\and
\small{$^{9}$Department of Electrical and Systems Engineering, University of Pennsylvania, Philadelphia, 19104, USA}\and
\small{$^{10}$Department of Psychology, Temple University, Philadelphia, 19122, USA}
\and
\small$^\ast$Corresponding author. Email: feifeiqi@usc.edu\and
\small$^\dagger$These authors contributed equally to this work.
}

\begin{document} 

\maketitle

\begin{abstract} \bfseries \boldmath
The ability to efficiently and effectively explore planetary surfaces is currently limited by the capability of wheeled rovers to traverse challenging terrains, and by pre-programmed data acquisition plans with limited in-situ flexibility. In this paper, we present two novel approaches to address these limitations: (i) high-mobility legged robots that use direct surface interactions to collect rich information about the terrain's mechanics to guide exploration; (ii) human-inspired data acquisition algorithms that enable robots to reason about scientific hypotheses and adapt exploration priorities based on incoming ground-sensing measurements. 
We successfully verify our approach through lab work and field deployments in two planetary analog environments. The new capability for legged robots to measure soil mechanical properties is shown to enable effective traversal of challenging terrains. When coupled with other geologic properties (\eg composition, thermal properties, and grain size data etc), soil mechanical measurements reveal key factors governing the formation and development of geologic environments. We then demonstrate how human-inspired algorithms turn terrain-sensing robots into teammates, by supporting more flexible and adaptive data collection decisions with human scientists. 
Our approach therefore enables exploration of a wider range of planetary environments and new substrate investigation opportunities through integrated human-robot systems that support maximum scientific return. 
\end{abstract}

\noindent
Humans have highly evolved traits that allow us to effectively explore and learn about complex environments. Legs allow us to traverse rough, steep and loose terrain~\cite{raibert1986legged,saranli2001rhex,playter2006bigdog,li2009sensitive,qian2013walking,qian2015principles,qian2015dynamics,bledt2018cheetah,qian2019obstacle,hutter2020learning}, and cognition lets us execute an adaptive (rather than fixed) search strategy to efficiently forage for information \cite{wilson2020,liu2023understanding,liu2024modelling}. 
Planetary surface exploration, however, remains largely conducted through wheeled robotic proxies, with activity sequences planned and uplinked (transmitted) from Earth \cite{gaines2016, milkovich2022percy, sun2024percy}, and with limited capacity for short-term, flexible and adaptive decision-making.

The properties of planetary regolith play a critical role in the mobility, surface interaction, and operational capabilities of robotic missions. Many planetary surfaces commonly display loose or unconsolidated regolith (Fig. \ref{Fig.fieldwork}A, B), including environments on the Moon, Mars, Titan, and Venus~\cite{vaughan2023percy, ewing2017, noble2009lunar, moroz1983venera, keller2008titan, clarke2009}. Such terrains pose significant challenges for the mobility of conventional wheeled rovers, with dangers of slippage, sinking, and entrapment~\cite{arvidson2017mars,li2008characterization} (Fig. \ref{Fig.fieldwork}A, B insets).
Furthermore, small changes in regolith properties, such as packing fraction~\cite{li2009sensitive,marvi2014sidewinding,qian2015principles}, chemical composition, grain size~\cite{sullivan2011cohesions, arvidson2017curiosity}, inclination~\cite{hutter2022traversing,liao2025bio}, and moisture content~\cite{liu2023adaptation,liu2025adaptive}, can cause large variations in mechanical responses during robot interaction. 
Because the mechanical and physical properties of planetary regolith are often uncertain, these factors collectively represent substantial risk for robotic navigation and measurement strategies, potentially resulting in lost mission time, missed scientific opportunities, or even in mission losses~\cite{sorice2021insight, maimone2007MER, callas2015mars, arvidson2017curiosity}.

Even the most recent generations of NASA Mars rover missions still face significant constraints in optimizing operations and maximizing science return~\cite{gaines2016, milkovich2022percy, vasavada2022mission, nachon2024percy, sun2024percy}). 
Operations demand substantial effort in planning, coordination, and the development of command products, all of which must be reactive to the downlink (reception) on Earth of the data resulting from activities executed by the rover during the previous sol(s). 
Science teams must rapidly assimilate multiscale datasets and identify high-value targets for subsequent uplinked activities. 
Additional challenges arise from communication limitations (including the intrinsic offset between Earth days and Mars sols), which can lead to insufficient data downlink for effective activity planning.
Collectively, these factors can constrain productivity, and lead to under-utilization of rover resources~\cite{gaines2016, sun2024percy}. 
In some cases, new scientific insights might emerge only after the rover has already driven away from a site of interest. 

Overall, a central challenge for planetary surface exploration remains to optimize time and resources while minimizing risks. 
These requirements for safety and efficiency are expected to grow as future missions deploy increasingly diverse and autonomous exploration platforms (e.g. as human-robots teams and multi-robot systems). They are especially relevant for the next key steps in the U.S. planetary exploration program \cite{SPD2017}, including the return of humans to the Moon, through the Artemis program \cite{whittington2025}, followed by future crewed missions to Mars \cite{merancy2024, NationalAcademies2025}.
On such upcoming crewed missions to the Moon and Mars, mobile robots could scout ahead of astronauts to identify safe routes and science rich targets, as well as work alongside astronauts to support \textit{in-situ} data collection decisions and flexibly adapt their locomotion and measurement strategies in response to changes in complex terrains. 

Our multi-disciplinary research team has developed an approach to meet this opportunity: \textbf{Legged Autonomous Surface Science in Analogue Environments (LASSIE)}.
LASSIE integrates legged robot mobility, proprioceptive sensing, and human-robot collaborative reasoning, to create next-generation environment-aware, discovery-driven platforms (Fig. \ref{Fig.overview}). 

\subsection*{Background and Rationale for LASSIE}

Recent developments in robotics have enabled the creation of high-mobility legged robots, inspired by the agile movement capabilities of various animals in nature. 
Legged robots offer unprecedented locomotion capabilities~\cite{roberts2014desert1,roberts2014desert2,roberts2019mitigating,hutter2015dynamic,hutter2020learning,hutter2022traversing} on many Earth and planetary surfaces; the higher degree-of-freedom in robot legs allows these platforms to utilize a versatile set of terrain-interaction strategies to negotiate steep and loose regolith.
The capability to successfully traverse these extreme terrains is essential for scientific explorations to locations with high scientific value that are not easily accessible by the wheeled rovers more commonly used in planetary operations.

A unique benefit of using legs is that direct surface interactions allow robots to collect rich information about the environment through every step (Fig.~\ref{Fig.overview}A). 
Robot legs, like the legs of humans and other tetrapods, are capable of ``feeling'' the surface they are moving over, sensing how it is moving or deforming underneath them. 
Recent advances in motor technology have allowed for lower-gear ratio motors~\cite{kenneally2016design} to be used in robotic joints while still achieving high performance. 
Quasi-direct-drive (with gear ratios lower than 10:1) and true-direct-drive (gearless) setups minimize gearing backlash within the joints of the robot, allowing for force transparency and thus enabling proprioceptive capabilities~\cite{kenneally2018actuator}. 
Within this paradigm of highly transparent robotic actuators, robots can reliably and precisely measure regolith force responses from their legs throughout the entire contact period with the surface.
To realize these opportunities and extend current planetary exploration, \textbf{LASSIE enables evaluation of a wider range of planetary-relevant environmental gradients and provides regolith strength measurements with high spatial and temporal resolution via a novel ``sensing through locomotion'' capability}.

For legged~\cite{saranli2001rhex,playter2006bigdog,qian2013walking, hutter2016anymal,bledt2018cheetah} and wheel-legged robots~\cite{chen2013quattroped,qian2015principles,chen2017turboquad} moving through complex, deformable regolith on Earth and other planetary surfaces, substrate forces “felt” through locomotive appendages at each step can be more informative~\cite{qian2019rapid,ruck2024downslope} than visual or other exteroceptive inputs for inferring environment properties and gradients, particularly on anisotropic regolith with varying levels of surface crust or ice content. 
These inferred surface properties can provide crucial guidance for robots to rapidly adapt locomotion strategies and avoid catastropic failures. 
Moreover, this proprioception-based ground sensing allows robots to collect rich information about the new environments that they are exploring through every step.
Such information is scientifically valuable for understanding the physics of landscapes and the geological processes that form them on other worlds, particularly when coupled with more traditional geoscience payload assessing thermal properties, composition, grain size, etc. from satellite and from robots on the surface (Fig.~\ref{Fig.overview}B). 
Moreover, sensing through locomotion comes at essentially no additional mass, power, or integration cost compared to payload instruments. 
Thus, \textbf{LASSIE contributes a new, high density, complementary source of surface information that enables convergent constraints on regolith properties without competing for payload resources or operational time.}

The increased density of data afforded by sensing through locomotion has the potential to alter how human scientists forage for information, necessitating a framework for developing and assessing workflows where human decision making is supported.
Therefore, \textbf{LASSIE couples the greater breadth of sampling opportunities with a shared autonomy framework for measurement decisions that enables robots to assist human teammates with decisions of where to collect data and how much to collect.}
The shared autonomy framework allows robots to independently identify scientifically valuable measurement targets using algorithms inspired by how expert scientists make dynamic data collection decisions \cite{liu2023understanding,liu2024modelling} (Fig.~\ref{Fig.overview}C).
These algorithms are distinct from traditional information theory based methods of robotic information gathering \cite{singh2007efficient, ghaffari2019sampling, pulido2020kriging}, where the alignment between autonomous robot decisions and human scientist decision processes is unclear. 
Instead, we take the perspective that human reasoning for scientific data collection problems should serve as a top-down guide for robot information gathering algorithm development. 

Our cognitively-informed approach has several key advantages over more traditional information theory methods.
First and foremost, it allows the robot to have a representation of thought processes, \ie theory of mind \cite{hiatt2011accommodating}, that lead to scientists' data acquisition preferences. 
This means the robot can infer scientists' underlying objectives from preferences, generate models of how objectives change over the course of data collection, and appropriately balance objectives in its decision making (Fig.~\ref{Fig.overview}C).
As a result, the robot has improved capability for explainable reasoning and use as a decision support agent: it can identify pros and cons of different potential data collection actions (in terms of objective fulfillment), and situations where adapting objectives (switching algorithms) might be needed based on a theoretically optimal model.
Also, scientists can provide explicit feedback on the robot's mapping of data collection objectives to choice preferences, making it easier to refine a model over time and without the extensive parameter tuning characteristic of traditional approaches. 
All together, this has the potential to make the robot a more intelligent and trustworthy teammate \cite{edmonds2019tale}, capable of making data collection decisions that allow for greater measurement flexibility in response to new observations and reduce the cognitive demand on scientists, thereby enhancing science outcomes. 

We present the LASSIE approach with a dynamically walking quadruped, demonstrating how the proprioceptive terrain sensing capability and human-inspired data acquisition algorithms can improve exploration in planetary missions operations (Movie S1). 
We verified our approach through lab work and through conducting field campaigns in two distinct planetary analogues: a dune field in White Sands National Park (New Mexico) that is an analogue for Mars' sand-rich and dune environments; and a volcanic ice-rich environment on Mt. Hood (Oregon) that is an analogue to lunar cratered icy landscapes, martian icy mid- and high-latitude areas, and surfaces of icy planetary bodies such as asteroids.
To achieve effective analogue campaign deployments we designed and validated novel leg sensing protocols and gaits that capture fine-scale regolith mechanical properties. 
Data and interpretations about regolith mechanical properties from the quadruped are validated by a standalone rheometer (single robot leg)~\cite{qian2017ground,ruck2024downslope}.
By deploying the legged robots alongside co-located geoscience instruments, we demonstrated how regolith mechanics measurements can inform about additional key physico-chemical properties of a planetary regolith and reveal processes that govern surface formation and evolution. 
Additionally, we created and tested the shared autonomy framework for rapid, adaptive in-situ data collection decision making in response to incoming information and shifting scientist objectives.
Our framework supports real-time field data collection decisions during campaigns, and the capability for explainable reasoning allowed scientists to identify and correct misalignment in human-robot objectives.
By comparing field observations of regolith mechanics to carefully controlled laboratory experiments in synthetic soils, we were able to discover how qualitative and quantitative variations in regolith strength arise from distinct material properties.

The success of the LASSIE project in planetary analogue environments demonstrates that our approach can enable legged robots to team with humans in a manner that facilitates maximum science return for upcoming and future planetary exploration missions, \eg on the Moon and Mars:
investigating new environmental gradients that are unreachable using wheeled rover systems, 
using proprioceptive sensing to provide regolith mechanical measurements with unrivaled spatial and temporal resolution (making every robotic step count),
and supporting integrated human-robot data collection decision-making for the advancement of exploration and scientific investigation of planetary bodies.

\section*{Results}\label{sec:results}

The primary goals of performing LASSIE fieldwork campaigns were two fold. 
First, to verify the potential for robotic legs to be used as scientific sensors. Second, to field test our human-robot shared autonomy workflow for decisions about where to acquire additional measurements in a disciplinary science campaign. 
To achieve the first goal, a quadrupedal robot and a standalone robotic leg were used to evaluate the force profile produced by the robotic legs interacting with the substrate (Fig. \ref{Fig.7}, A-C) \cite{methods}. 
The standalone leg removes dynamic and complex forces associated with walking and provides a ``ground truth'' of regolith mechanics~\cite{qian2019rapid,ruck2024downslope}, whereas the quadruped rapidly collects spatially dense data during locomotion, complementing the localized measurements. 
In addition, a battery of co-located, complementary measurements was collected by geoscientists to characterize surface and shallow subsurface properties (Fig. \ref{Fig.7}, E, G, I, J) \cite{methods}. 
This suite of geoscience measurements included several analytical techniques analogous to those deployed on space missions, including onboard the two most recent generations of NASA Mars rovers. 
To accomplish the second goal, geoscientists were given suggestions from the robot (via a cognitively-inspired algorithm, run on a tablet interface) about where to acquire force measurements next, and were interviewed by cognitive scientists in the field about their decisions to accept or reject the robot's suggestions (Fig.~\ref{Fig.Ops}A).  

The fieldwork campaigns were carried out at two distinct planetary analogue sites --- White Sands, NM (Fig.~\ref{Fig.fieldwork}F-J), and Mt. Hood, OR (Fig.~\ref{Fig.fieldwork}K-N). 
These sites span a range of surface textures and depositional contexts, providing natural gradients for evaluating in-situ measurements of regolith mechanical properties. 
Measurements were performed at several spatial scales, ranging from \textit{Regions}, to \textit{Transects}, to \textit{Locations}. Regions range from kilometric to decametric in size, and were selected by geoscientists to be representative of the sites' diversity in landscape features and surface textures.
In each Region, geoscientists identified key Transects that captured a gradient of surface textures, which ranged from 5 to 150 meters long depending on the variability of surface textures observed.

\subsection*{Field Campaign 1: White Sands National Park (New Mexico), A Dune Field Planetary Analogue }\label{sec:results:sites-WS}

White Sands is an active gypsum dune field in a playa-lake setting that displays a variety of sedimentary features and surface textures, including loose sand, partially lithified sandstones, and various salt- and microbial-crusts. 
Our Region of interest comprised isolated barchan dunes of loose and mobile gypsum sand migrating over seasonally wet gypsum sand due to the shallow groundwater table \cite{jerolmack2012internal,ewing2012morphology,ewing2020white}. 
This site provides an analogue for sedimentary environments on Mars, e.g. ancient playa-lake deposits explored at Meridiani Planum with the NASA Opportunity rover \cite{grotzinger2005stratigraphy, chavdarian2006cracks, jerolmack2006spatial}, and the gypsum dunes of the high Mars latitudes (observed with satellite images). 

The key Transect was located in between two barchan dunes (Fig.~\ref{Fig.WS}A), providing a smooth spatial gradient of surface textures (Fig. \ref{Fig.WS}B). 
Barchan dunes consist of an erosional windward (stoss) slope and a steeper depositional downwind (lee) slope, while interdune regions are relatively flat,  with water table closer to the surface, and thus may host abiotic salt crust or biotic crusts formed by vegetation and/or microbes~\cite{jerolmack2012internal,qian2019rapid, ewing2020white, reitz2010barchan, lee2019imprint}.
Although White Sands is an active dune field, the dunes' relatively slow overall migration (meters per year) occurs on a timescale that did not affect our measurements of the surface and shallow subsurface.

The environmental features of a dune field present unique challenges in robotic locomotion, sensing, and operation planning. 
First, moving through highly deformable sand can be extremely challenging for wheeled rovers.
Sandy surfaces are often visually homogeneous, and therefore it can be difficult to remotely identify safe regions \emph{a priori}, based exclusively on visual data.
While some of the terramechanics can be estimated or deduced from the shape of the dunes and the predominant wind direction that mobilizes sand, the legged robots' ability to provide proprioceptive, tactile feedback while being general and capable enough to avoid getting stuck, could allow scientists to explore and test hypotheses in locations otherwise inaccessible to wheeled rovers.

A second challenge is resolving subtle but scientifically meaningful variations in regolith mechanical properties. 
At White Sands, biological activity (vegetation, micro-organism activity in soil) increases along the direction of wind direction and dune stabilization~\cite{jerolmack2012internal, qian2019rapid, ewing2020white, reitz2010barchan, lee2019imprint}, potentially altering substrate response during robot–ground interaction.
Disentangling biomechanical signatures from purely physical effects raises the possibility that legged robots could identify biologically influenced substrates through locomotion-based sensing. 

A final challenge is how to forage for information in environments where every step can provide scientifically valuable data. 
Even if proprioceptive measurements are low-cost and continuous, the decision space for how to use them is almost infinite: a legged robot can locomote in many possible directions and deploy multiple different sensing protocols through locomotion. 
With this increase in degrees of freedom, the difficulty of deciding which next action will most improve scientific understanding also increases.
Developing methods that can adapt exploration strategies in real-time, as new measurements update the scientific hypothesis, could allow robots and scientists to rapidly converge on the most informative locations and discover phenomena that would otherwise remain undetected. 

To address these challenges, we deployed both the quadruped and the standalone robotic leg, coupled with the localized acquisition of a suite of surface and shallow subsurface geoscience measurements (including surface topography, compositional analysis, and grain size measurements), within a human-robot shared autonomy workflow. 
Together, our deployment demonstrates how robotic legs can serve as \textit{in-situ} scientific instrumentation, providing densely distributed force measurements that help reveal key geological properties of regolith that govern underlying crust formation and dune stabilization.

Figure \ref{Fig.WS} presents measurements from our field deployment at White Sands National Park and shows the selected Transect (Fig. \ref{Fig.WS}A-B) along the dominant wind direction, spanning from the lee face of one dune, across the interdune area, and onto the stoss face of the adjacent dune. 
Surface texture imagery along this transect (Fig. \ref{Fig.WS}C) reveals a progression from smooth, unconsolidated regolith (Location 0) to increasingly encrusted surfaces with higher roughness and fractured layering (Location 5, 15). 
Relating such surface textures to underlying mechanical behavior under environmental forcing (\eg wind erosion) is central to understanding landscape evolution, yet remains difficult to infer from imagery alone. 

The proprioceptive force measurements by the robotic leg offer direct access to characterizing mechanical variability of the shallow subsurface, providing key rheological and stratigraphic information that complements visual observations. 
Across the transect, penetration resistance force consistently increased moving from the dune lee face through the interdune (Fig.~\ref{Fig.WS}D).
At the dune slip face (Location 0, Fig.~\ref{Fig.WS}D, 0m), unconsolidated sand exhibited low resistance ($<$5 N) and allowed deep penetration (up to 4 cm). 
At lightly encrusted interdune locations (Location 5, Fig.~\ref{Fig.WS}D, 5m), resistance increased rapidly, reaching $\sim$15 N at 2 cm depth and limiting penetration, with distinctive force drop when surface crust ruptured under applied stress.
Further into the interdune (Location 15, Fig.~\ref{Fig.WS}D, 33m), forces rose sharply to $\sim$30 N at $\sim$1 cm depth, indicating well-developed crusts and solidified strata associated with repeated dune migration.
Regolith strength decreased again approaching the stoss face of the next dune.
These trends are consistent with existing geomorphological models of dune stabilization at White Sands~\cite{jerolmack2012internal, qian2019rapid, gunn2020macroscopic}, providing in-situ data to validate theoretical models of sediment transport and dune morphology evolution. 
This integrated approach directly addresses the robotic sensing challenge discussed above: enabling robotic legs to function as scientific instrumentation that reveal the dominant physical processes shaping the environment. 
By linking mechanical, morphological, and compositional data, the robot provides key evidence to advance our understanding of insights into the formation, evolution, and stabilization of landscapes. 

These mechanical profiles are also directly linked to the variability in composition and grain size observed from our trenching experiments (Fig. \ref{Fig.WS}E), enabling identification of key sedimentary layers, particularly the structure and strength of surface crusts, and their roles in sediment transport and dune migration. 
The robotic leg reliably distinguished weak, unconsolidated sand from hardened surface crusts overlaying looser material beneath. Such distinctions, which are often indistinguishable from surface appearance alone, highlight the importance of force-based measurements for both safe mobility and scientific exploration.
These results directly addresses the robotics locomotion challenge discussed above: enabling legged robots to support the safe traversal of wheeled rovers and access regions that were previously deemed too risky due to uncertain terrain mechanics.

The new capability of using legs as regolith sensors also opens up exciting avenues for operation planning. 
With robotic legs' capability as in situ science instruments, they can be leveraged to identify areas of scientific interest and assess traversal risk in real time, enhancing both mission planning and scientific discovery. 

Figure \ref{Fig.Ops}B shows the shared autonomy workflow enables robots to assist humans with decisions about where to collect data next, and supports the discovery of new data collection objectives and strategies \cite{methods}. 
In three Regions at White Sands, with three different geoscientists, data on regolith strength from an initial set of scientist-selected locations along a Transect were used by the robot to generate multiple rounds of suggestions about which Location to collect data next along the Transect (see \ref{Fig.Ops}B-i). 
Suggestions were generated based on the geoscientists' objective to evaluate the hypothesis that regolith strength increased moving from the lee face of one dune, across the interdune, to the slip face of the adjacent dune. 

All geoscientists accepted at least one suggested Location from the multiple suggestions (2 to 4 rounds each), with an overall acceptance rate of ~70\% (6 of 9 total rounds). 
Interestingly, the chosen suggested Locations were never the ones that were maximally rewarding for the algorithmic representations of the objectives (see Fig.~\ref{Fig.Ops}B-ii). 
Instead, the geoscientists chose suggested Locations that balanced hypothesis testing with broader information gathering, (Fig. \ref{Fig.Ops}B-iii), and adjusted this balance as data accumulated. 
This reveals that fixed, single-weight models of scientific objectives do not capture how experts actually reason. 
Scientists dynamically adjust how they trade off objectives as information accumulates and hypotheses evolve.
This result provides an opportunity to improve robotic data collection algorithms to incorporate dynamic, data-dependent shifts in objective weighting, supporting the operational challenges described above: enabling legged robots to identify scientifically-interesting sites and rapidly gather information about a new environment.

In summary, results from the White Sands Analogue Field Campaign show that robotic legs can help identify geotechnically and scientifically relevant areas through force sensing. 
By combining with the human-inspired autonomy, our approach enables human-robot teaming to support measurement of high-science value locations, on the way to achieving rapid and reactive adaptation of robot operations for scientific exploration and mobility-driven scouting.  

\subsection*{Field Campaign 2: Mount Hood (Oregon), An Icy Volcanic Planetary Analogue}\label{sec:results:sites-MH}

Mount Hood is a stratovolcano, located along the Cascade Mountain Range, and consists mainly of pyroclastic and debris flows as well as glacial till (unsorted and un-stratified material deposited by glacial ice) \cite{crandell1980, geolmapmthood}. 
Such icy volcanic landscapes exhibit natural gradients in ice content, water content, and particle composition, providing a testbed for investigating the mechanical behavior of frozen and partially frozen soils.  
The unconsolidated volcanic sediments on Mount Hood’s slopes, combined with the presence of surface and subsurface ice, form a relevant analogue for icy regolith in lunar polar craters, martian mid- to high-latitude terrains, and other ice-rich planetary bodies.

Our Region of interest at Mount Hood included slopes of varying steepness composed of loose volcaniclastic material, spanning grain sizes from silt to boulders, interspersed with irregular patches of surface ice.
Compared to the White Sands campaign, this environment introduced additional operational and sensing challenges beyond differences in lithology and rheology.
Most notably, transitions between bare regolith and ice occurred abruptly along irregular contours controlled by local topography (Fig.~\ref{Fig.MH}A).
This pronounced spatial heterogeneity, or ``patchiness'', necessitated a different data collection strategy: rather than a single long transect capturing a smooth gradient, the scientific questions required measurements that resolved micro-scale transitions across multiple ice–regolith boundaries.
Such measurements are essential for probing how proximity to surface ice influences shallow subsurface mechanics, including the onset and depth of partial freezing.
Developing methods that can adapt exploration strategies to patchy environments would enable robots and scientists to allocate sensing precision where it is most informative, improving the efficiency of operations and maximizing science return.  

The second key challenge is relating force-based measurements to the texture and formation mechanisms of icy regolith. 
Icy regolith is of high scientific interest for planetary exploration \cite{li2018direct}, but the physical form of these ice-soil mixtures, and their link to past environmental processes, remain poorly understood \cite{moore2014deformation}. 
Ice can be incorporated into regolith in multiple ways, such as cemented layers, surface frosts, granular ice particles, or vapor deposition~\cite{ricardo2023review}. 
Each of these forms could imply a distinct scientific hypothesis of how water was delivered or cycled on the Moon. 
We hypothesized that these different modes of formation would produce distinct mechanical behaviors. 
Thus, by measuring the force responses of icy regolith, we could begin to address key questions about the origin and distribution of lunar water, and whether analogous processes could occur on other planetary bodies that are of interest for future human habitation.

To address these challenges, we deployed our legged robotic setup, coupled with a geosciences suite of measurements for characterizing the surface and shallow-subsurface, including thermal images via drone, compositional measurements, macro-images of surface textures, and water-content data). 
This environment required a new data collection workflow: rather than laying a single long Transect to capture a dominant spatial gradient, geoscientists first identified multiple Regions of Interest (ROI) where local transitions were expected, and then laid multiple short Transects to resolve micro-gradients across these boundaries (Fig. \ref{Fig.Ops}C). 
We used a refined version of the human-robot shared autonomy workflow that incorporated explanations of the multi-objective reward information~\cite{liu2024modelling} the robot used to make its suggestions of Locations to acquire measurements along the Transect. 
Suggestions were generated based on the geoscientists objective to evaluate the hypothesis that regolith strength would be would be greatest in the soil-rich ice at the boundary of ice patch, and diminish with the transition into ice-free soil.

When deployed with a geoscientist, the explanation of the robot’s suggestions was reported to be overall helpful for understanding how decisions were generated \cite{flynn2024>}, and allowed for identification of a mismatch in representation of the information objective between the robot and the geoscientist. 
Specifically, the robot used an information gathering objective learned from expert data collection behaviors in environments with smooth spatial gradients, where the value of information is approximately uniform across the gradient.
In contrast, the geoscientist was focused on micro-gradients in transitions between identified ROI, where the value of information is highest near ROI boundaries, and decreases rapidly towards the ROI center. As a result, relatively few measurements were needed within homogeneous interior regions. 

Discovery of this new representation of information for ``patchy'' environments presents a compelling use case for legged robots' capability to dynamically modulate gait and sensing precision. For example, robots can deploy slow, high-precision behaviors (\eg \crawlgait{}) near ROI boundaries while using faster gaits (\eg \trotgait{}) to move through homogeneous regions between boundaries. 
This directly addresses the operations challenge described above: enabling legged robots to identify \textit{where} high science value areas are and \textit{how} to gather information efficiently.

Based on the data collection behavior observed from patchy environments, we deployed the quadrupedal robot to sample along the short Transects across ROI boundaries. 
Figure \ref{Fig.MH}A and B shows the aerial and ground view of our analog site, overlaid with tracks from the robot’s walking path (Fig. \ref{Fig.MH}A, circular markers). By employing the \crawlgait{}, the legged robot produced a high-resolution 2D map of regolith strength (Fig. \ref{Fig.MH}C). The spatial distribution of robot-measured regolith strength closely corresponds to the soil-snow boundaries visible in the aerial image (\eg Fig. \ref{Fig.MH}A, Path 4), illustrating the rapid variations of substrate properties spatially.

Figure \ref{Fig.MH}G-I present the force–depth profiles (bottom panels) at three of these representative scientific target locations, alongside their corresponding macro-imaging (top panels). The force data reveal distinct rupture signatures (\ie sudden drops in resistance) that correspond to transitions in ice content and structure. For regolith with relatively low ice content (Fig. \ref{Fig.MH}I), the force-depth curve exhibit a linear trend, similar to dry granular material, where force responses are dominated by interparticle friction and particle rearrangements. For regolith with higher ice content, interstitial frozen liquid begins to bridge grains and provide cohesive strength, and the substrate begins to yield as a ‘ductile’ material marked by an elastic loading regime and followed by an elongated plastic region from 1.5 to 3.5 cm deep (Fig. \ref{Fig.MH}H). In surfaces covered in a loosely packed snow (Fig. \ref{Fig.MH}G), the overall resistance decreased significantly, while exhibiting repeated ‘brittle' failure as frozen liquid bridge ruptures. 

These mechanical behaviors are corroborated by geoscience images (Fig. \ref{Fig.MH}G-I top panels) and thermal observations (Fig. \ref{Fig.MH}E-F), enabling identification of scientifically valuable regions for targeted compositional, thermal, and microscopic analyses. For example, Figure \ref{Fig.MH}J-K shows a trench exposing the upper centimeters of a local subsurface along a transect from surface ice- and snow-covered ground to bare rocky regolith. This example illustrates that visual assessment of surface textures alone might be insufficient to infer subsurface properties. The force data collected by robotic legs at each step can provide informative cues about environment properties and gradients.

In addition, by enabling rapid, spatially resolved mapping that integrates proprioceptive and remote-sensing data, the LASSIE approach allows capturing of transient environmental states that would otherwise be missed in traditional field or orbital observations. Figure \ref{Fig.MH}E-F shows thermal drone views of this field site site, at two different times. Comparison between these panels reveals pronounced temporal variability: within only a few hours, the extent of the frozen regolith patch visibly contracted as surface ice melted and redistributed. 

Overall, the robotic-leg capabilities for step-by-step sensing and adaptation are especially relevant for future planetary exploration of icy regolith, such as permanently shadowed lunar regions (PSRs) and martian mid-latitude icy terrains. Our results show that force-depth data from legged robots can provide in situ characterization of icy regolith, yielding critical information to test hypotheses about volatile geologic history and to evaluate the accessibility of potential resources for in-situ resource utilization (ISRU). Importantly, rupture events also coincided with observed locomotion failures, highlighting the dual role of proprioceptive detection: advancing science while ensuring safe mobility on icy terrain. 

\subsection*{Laboratory Results: Isolating Controls on Regolith Strength to Explain Field Observations}
\label{sec:results:lab}

Observing what nature has created guarantees ``field relevance'', but extrapolating from these observations is dangerous; we cannot isolate variables to definitively test our hypotheses, and we cannot know whether we have observed the most important factors. Laboratory experiments are a complementary approach: building and varying synthetic regolith one component at a time, to identify the minimal set of ingredients and conditions that are required to produce the mechanical behaviors observed in nature \cite{pradeep2025geomimicry}.    
More, we can test the sensitivity of regolith sensing protocols under controlled conditions, to determine whether our field methods are sufficiently robust. While the mechanical behavior of regolith and its sensitivity to material changes has its own intrinsic scientific interest, we focus here on the portions of laboratory work that directly support LASSIE activities. 

For the purposes of mapping regolith properties on a landscape, and for developing a terrain mechanics model that can be used for locomotion, a simplified description of regolith ``strength'' is desired. Locomotion studies often parameterize regolith force response to vertical intrusion, $F_z$, as a one-way spring \cite{zhao, li2009sensitive,zhang2014effectiveness,qian2015principles}; i.e., $F_z = kh$, where $h$ is penetration depth and $k$ is the penetration resistance (\ie the ``stiffness'' of the one-way spring). We know of course that real granular materials are not springs; the question is whether and how such a simplified approach can be justified, and what material property is described by the parameter $k$. Granular physics studies have revealed that the penetration resistance of sand is determined by the volume of displaced material, $V = V_0 +hA$, where $V_0$ is the volume of a stagnant cone of grains that forms under the intruder at the internal friction angle, and $A$ is cross-sectional area of the intruder \cite{kang}. Thus, $F_z = K \phi\rho_pgV$, where $\phi$, $\rho_p$ and $g$ are particle volume fraction, particle density and gravity, respectively. In this formulation, $K$ is a constitutive property of the material that depends only on granular friction, $\mu$ \cite{brzinski, kang}. This ``Archimedes'' model, which strictly applies only to slow (quasi-static) intrusion into cohesionless particles, maps the ``one-way spring'' parameterization to a physically interpretable material property, while also suggesting limits to its applicability. 

We test the Archimedes model in the laboratory using a custom air-fluidizing bed, which allows us to create reproducible granular packs of monodisperse quartz sand at specified values for $\phi$. We use the same robotic standalone leg as the field to perform identical penetration tests (Fig. \ref{RuckDougFig}A). We varied penetration speed and intruder tip geometry, and found little to no effect on the resulting force curves \cite{ruck2024downslope} -- as expected from theory. We further performed penetration tests in differently prepared (cohesionless) granular beds, in materials with varying particle sizes and shapes -- including lunar simulant LHS-1 with the cohesive dust component removed. After an initial transient regime associated with formation of the stagnant cone, all granular data showed a linear increase of force with depth (Fig. \ref{RuckDougFig}C). For cohesionless sand, it is well known that the volume fraction $\phi$ is the primary variable controlling friction \cite{schroter,liu_song_ling,schofield1968critical}. 
We systematically varied $\phi$ from the loosest to the tightest possible packing, and found the same nonlinear increase in $K$ with $\phi$ for all materials examined (Fig. \ref{RuckDougFig}C) \cite{Ruck2025Unified}. More, the topographic deformation around the intruder systematically changed with $\phi$, suggesting that robotic footprints can provide information that corroborates and expands on force response data (Fig. \ref{RuckDougFig}C, inset). 

The results above confirm the validity of the Archimedes model for cohesionless regolith, indicating that we may infer meaningful material properties from field measurements. As seen above, however, many natural regolith materials we encountered are granular admixtures; sand plus some additional, often cohesive, component (Fig. \ref{RuckDougFig}B). To help understand the varied mechanical properties observed in field regolith, we created a few granular admixtures in the laboratory. Adding cohesive powder -- dust for LHS-1, and kaolin clay to our quartz sand -- affects a qualitative change in the force response that is well known for cohesive materials \cite{yang, trappe, Mandal2020CohesiveRheology}: the force response plateaus, rather than continuously increasing with depth, due to the presence of a yield stress. Adding liquid water to sand and then freezing it, to simulate icy regolith, produces force responses that are similar to curves observed at Mt. Hood. Ice elicits a brittle response; increasing ice content produces increasing strength, and also increasing intensity of stress drops as ice bridges among grains are broken (Fig. \ref{RuckDougFig}D). This response is distinct from moist (liquid water) and muddy regolith, which exhibits a smooth and nonlinear ``ductile'' force response. We have even experimented with salt crusts, formed by evaporating brine from a sandy bed. Similar to field observations at White Sands, salt crusts elicit a brittle response as the surface crust is punctured, followed by a typical sand response. The detailed mechanical interpretation of these experiments is beyond the scope of this paper, and will be presented elsewhere. For the purposes of LASSIE, our results suggest that linear force responses in the field indicate cohesionless materials and may be used quantitatively to determine granular friction. Deviations from linearity arise from distinct additions to sand -- such as cohesive grains, ice bridges or salt crusts -- that, with context such as compositional analysis, can be interpreted quantitatively. 

\section*{Discussion}

The LASSIE framework establishes a new mode of planetary surface investigation in which locomotion itself becomes a scientific measurement process. By integrating legged mobility, proprioceptive force sensing, and adaptive sampling strategies, our approach enables spatially dense mechanical characterization of regolith in dynamic and heterogeneous environments. Across two planetary analogue sites, we demonstrate that force–depth measurements collected during routine locomotion provide quantitative constraints on substrate strength, layering, and rheological state, with direct implications for sediment transport, crust formation, and volatile-related processes.

A central scientific contribution of this work is the identification of recurring mechanical signatures across physically distinct materials. For example, brittle rupture events were observed both in ice-rich regolith and in cemented sandy crusts, while extended plastic deformation regimes characterized both snow-dominated substrates and cohesive sandy mixtures. These parallels suggest that mechanical response functions can serve as a unifying framework for interpreting regolith state across environmental contexts. Rather than treating terrain mechanics solely as a mobility constraint, our results elevate force–depth behavior to a geophysical observable that informs hypotheses about compaction, cohesion, ice bonding, and subsurface structure.

Importantly, these measurements are not isolated point samples but form spatially continuous datasets. By embedding sensing within locomotion, LASSIE enables high-resolution mechanical mapping along transects and across regions, capturing gradients that would be impractical to resolve through conventional sampling alone. In dune environments, this approach constrains crust development and stabilization processes. In icy volcanic terrain, it distinguishes cohesionless, ductile, and brittle regimes associated with varying ice content. Such distinctions are directly relevant to planetary science questions regarding volatile distribution, regolith evolution, and landscape dynamics.

The broader implication is illustrated conceptually in Figure \ref{Fig.vision}. In this framework, legged locomotion produces continuous regolith mechanical measurements that are integrated with complementary payload data (e.g., thermal and compositional sensing). Detected transitions, such as changes in ice content, update scientific hypotheses, which in turn inform adaptive modifications to sensing path and gait. This closed-loop structure links measurement, interpretation, and action. Rather than separating reconnaissance, analysis, and planning into discrete phases, the system operates as an iterative hypothesis-testing engine in which each step refines understanding of the environment.

The integration of cognitively inspired decision models further strengthens this scientific workflow. By dynamically balancing information gathering and hypothesis verification objectives, the robot assists in prioritizing sampling locations in a manner aligned with expert reasoning. This capability is particularly important in ``patchy'' environments, where information value is concentrated near boundaries and transitions. The ability to modulate sensing precision and locomotion strategy in response to detected gradients allows efficient allocation of measurement effort while preserving spatial coverage.
Going forward, we envision that the integrated sensing gait, human-inspired autonomy, and principles to connect regolith properties to landscape formation, will enable fundamentally new science exploration modes (Fig. \ref{Fig.vision}):

First, with the technology developed from this paper, future robots could autonomously switch between different leg sensing protocols based on scientific hypotheses and locomotion efficiency. For example, robots could use a fast, locomotion-optimized gait to perform scout/recon, covering a wide range of planetary surfaces while measuring regolith mechanics. These high spatio-temporal resolution scouting measurements can enable identification of scientifically-interesting locations or spatial gradients for the mission. Once the scientifically-interesting locations are identified, the robot could flexibly switch to a sensing-optimized gait, to move along the scientific traverse, while performing high-precision force measurements every step along the way, detecting subtle regolith texture such as surface crust, subsurface ice, or existence of bio-polymer. 
These force measurements can be further integrated with complementary scientific sensor payloads, to inform the landscape formation process or past biological activities. 

Secondly, leveraging the cognitively-inspired algorithms, future platforms could flexibly switch between exploration and verification, to support hypotheses testing based on dynamic incoming measurements. For example, the robot could use the exploration-based rewards to identify a scouting path, and then switch to verification-based rewards if a sufficient discrepancy between measurements and hypotheses were observed. In addition, the robot could adopt different sensing gaits (\eg vertical penetration, horizontal shear, repeated stepping) to test specific hypotheses. These capabilities allows robots to assist beyond the level of sensor payload carrier, and become intelligent teammates that can connect science objectives to adapt exploration plans.

Lastly, these technology can be deployed into robotic-human hybrid team of legged-rover(s) and astronaut(s) to enhance scientific outcomes and mission efficiency. This approach is particularly relevant for the development of future missions, such as to Mars, where human-robot roles are expected to continue evolving depending on final mission architecture, science objectives and the duration and phases of surface operation \cite{Report2023, merancy2024, NationalAcademies2025}. The high-mobility, sensory legged robot could, for example, explore landing sites, assess their scientific value, and evaluate terrain risks. Such tasks could occur either during the crewed surface mission, or in a pre-crew phase, and could also continue after humans departure. This extended operational capability would extend the operational capacity of human crews, and increase the mission feasibility, especially in notional ``short-stay'' Mars missions scenarios featuring a 30-martian-days surface mission \cite{Report2023, NationalAcademies2025}.

These advances are particularly relevant for exploration of volatile-rich planetary environments, including permanently shadowed lunar regions and martian mid-latitude icy terrains. In such settings, mechanical signatures provide constraints on the physical state and accessibility of subsurface ice, informing both scientific interpretation and operational planning. By converting locomotion into distributed geotechnical probing, LASSIE demonstrates that high-density mechanical measurements can enhance both scientific discovery and mission resilience.

In summary, this work presents a unified framework in which legged robots act not only as mobility platforms but as active instruments for planetary surface science. By embedding mechanistic regolith probing within adaptive exploration workflows, LASSIE bridges robotics, granular physics, and planetary geoscience to enable a more responsive and information-rich mode of planetary exploration.


\begin{figure}
\centering %
\includegraphics[width=0.85\textwidth]{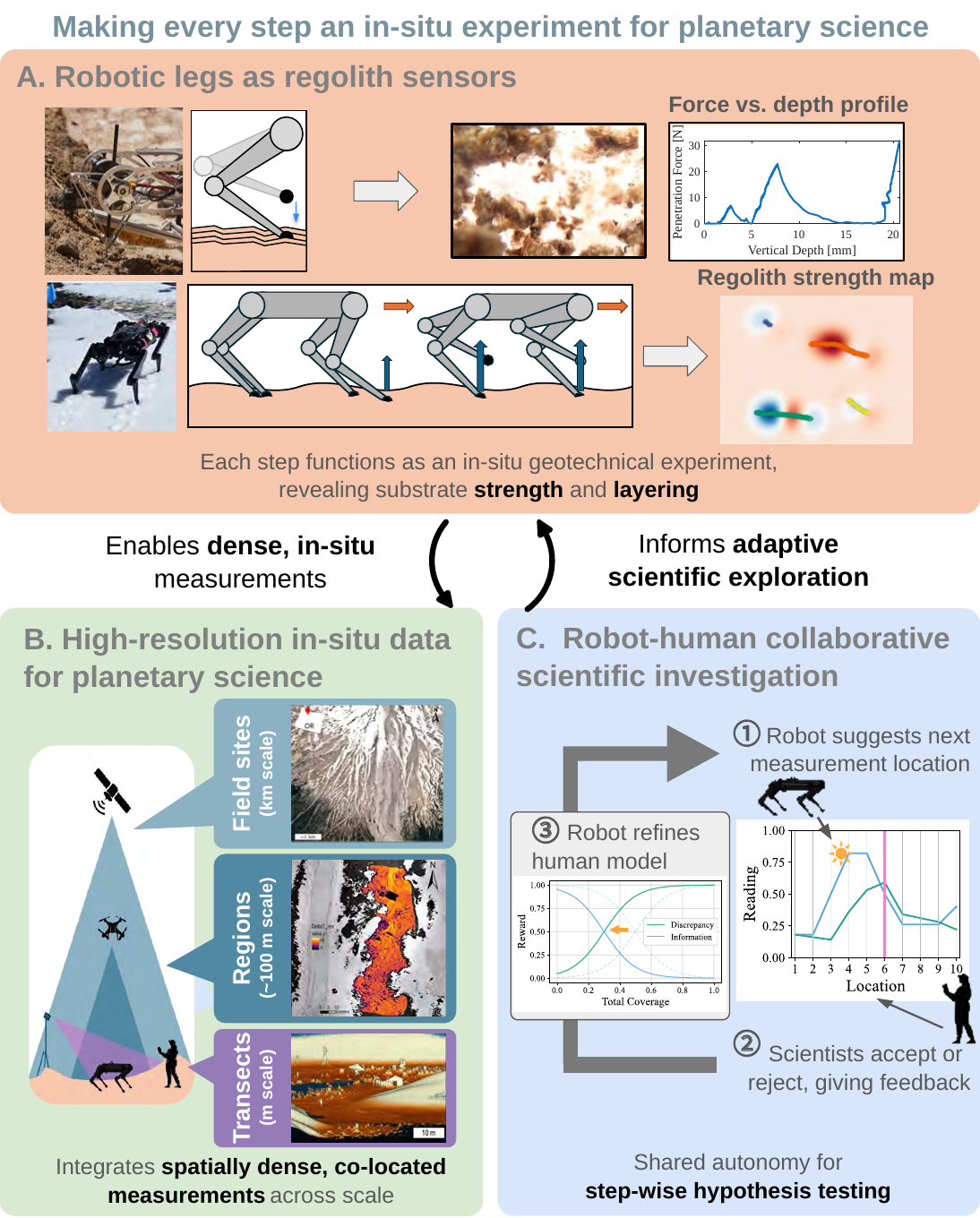}
\caption{\textbf{The LASSIE approach.} Using (\textbf{A}) integrated legged locomotion and sensing, (\textbf{B}) co-located geoscience measurements across scales, and (\textbf{C}) human-robot collaborative algorithms to enhance planetary science exploration.}
\label{Fig.overview}
\end{figure}

\begin{figure}
\centering
\includegraphics[width=0.85\textwidth]{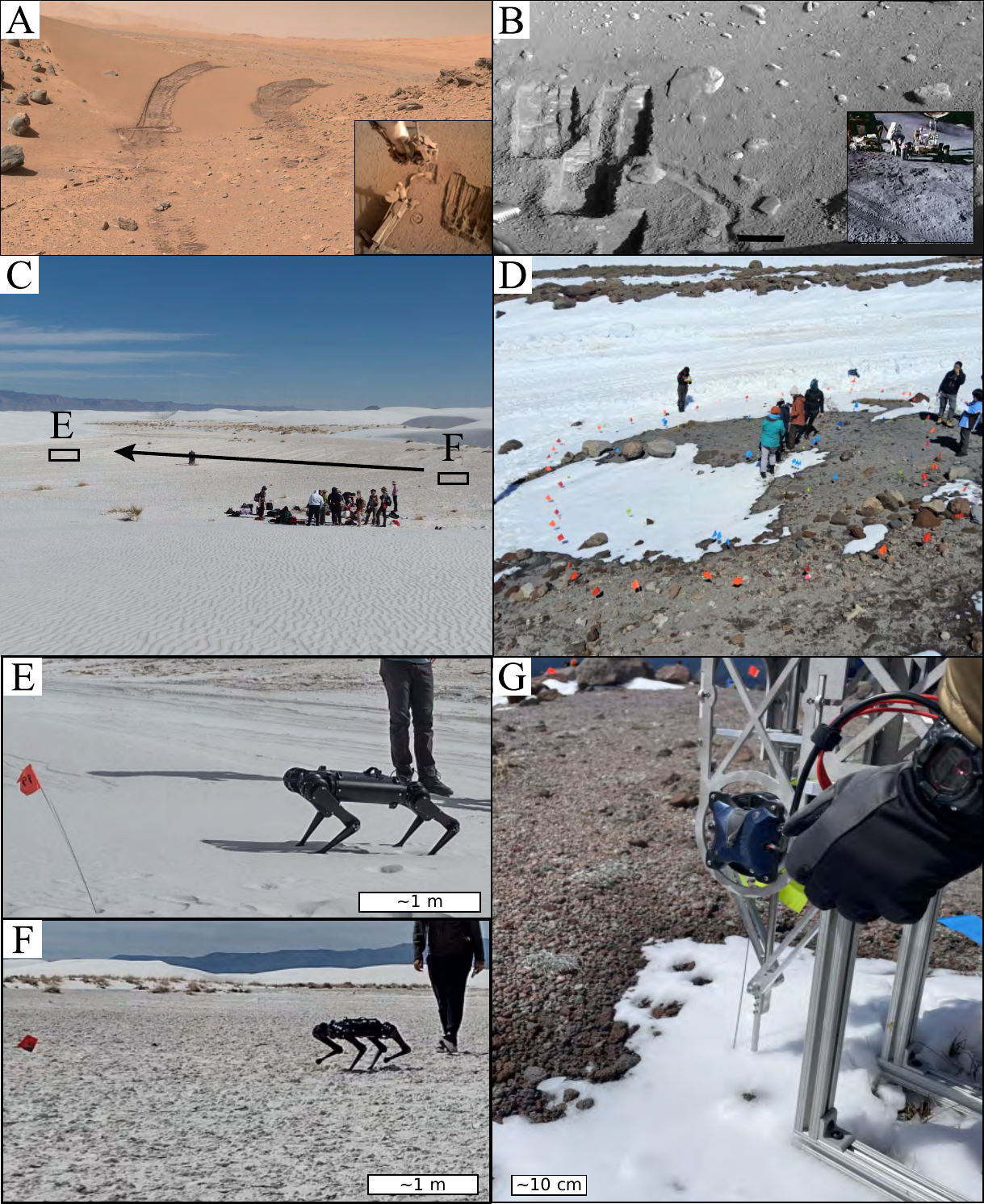} 
\caption{\textbf{Analogue field campaigns.} Analogues for (\textbf{A}) sandy and (\textbf{B}) icy environments on the Moon and Mars, where regolith mechanics critically affect robotic operations (insets), were, respectively, (\textbf{C}) White Sands National Park, NM and (\textbf{D}) Mt. Hood, OR. Representative regolith surface textures investigated at White Sands and Mount Hood are shown in (E–F) and (G). \small{\textit{Mars and Moon images IDs: CX00538ML0260438F445256775VA, D001R1239\_706529559EDR\_F0404\_0010M, s\_084eff\_cyl\_sr19c5e\_r111m1\_0PCT, AS15\-86\-11599.}}} 
\label{Fig.fieldwork}
\end{figure}

\begin{figure}
\centering
\includegraphics[width=0.85\textwidth]{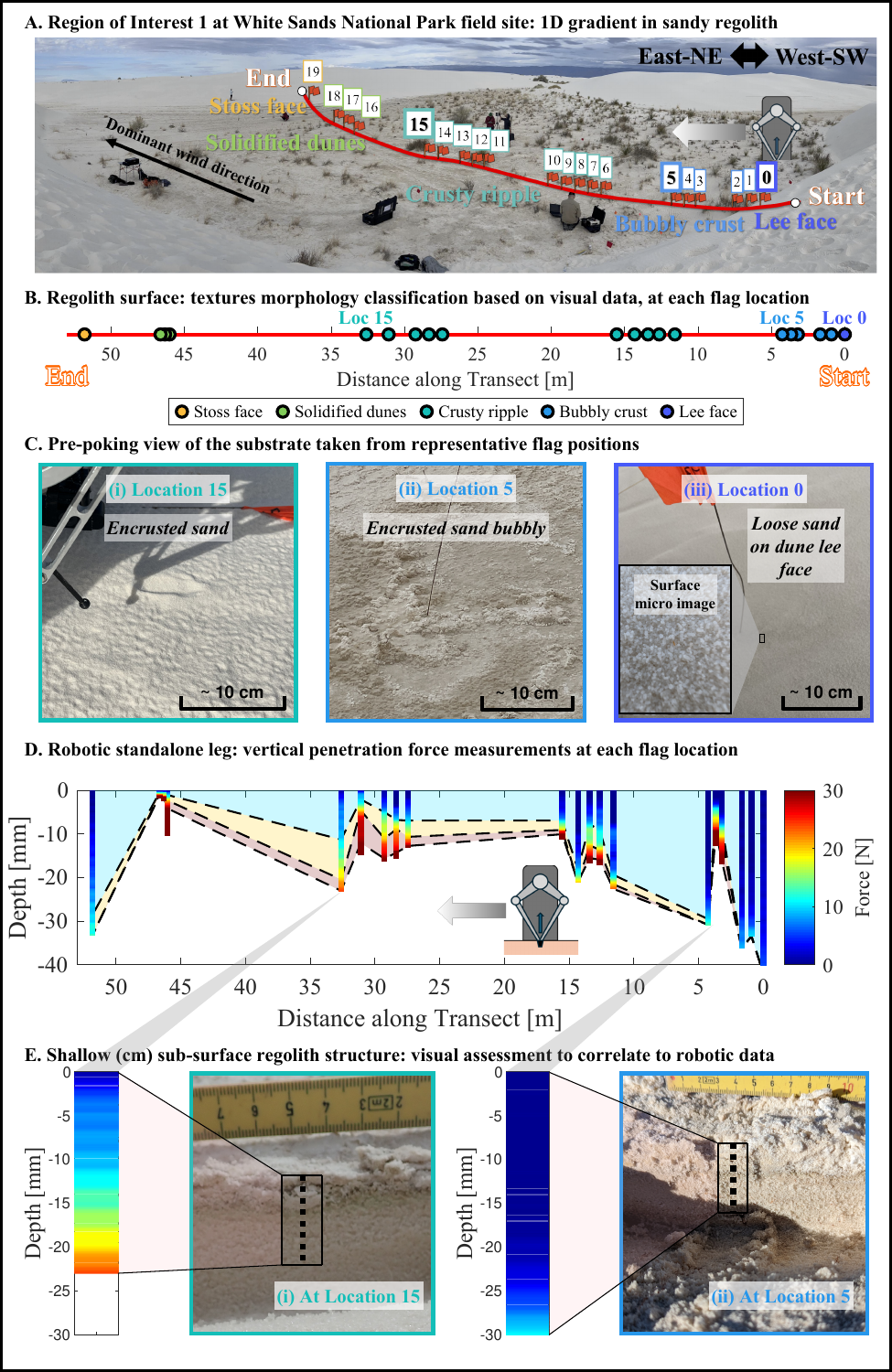} 
\caption{\textbf{White Sands field campaign results.} }
\label{Fig.WS}
\end{figure}
 \addtocounter{figure}{-1} 
\begin{figure}[t!]
    \centering
    \caption{(Continued from previous page.) Data were acquired along (\textbf{A}) a 55 m transect from a dune’s lee face, across the interdune, to the next dune’s stoss face, at (\textbf{B}) 20 geologist-selected and 4 robot-selected flagged locations. At each site, (\textbf{C}) surface morphology and composition were recorded prior to robot leg interaction. The (\textbf{D}) regolith resistive force along intrusion depth measured by the standalone robotic leg was shown as colored columns. Blue, yellow, and brown shaded regions represent the depth corresponding to 10, 20, and 30 N forces. We highlight (\textbf{E}) mini-trenching images (5–10 cm subsurface exposure) at Flags 5 and Flag 15 here, which correspond well with depth–force measured by the robotic leg.)
    }
\end{figure}

\begin{figure}
 \centering %
 \includegraphics[width=0.85\textwidth]{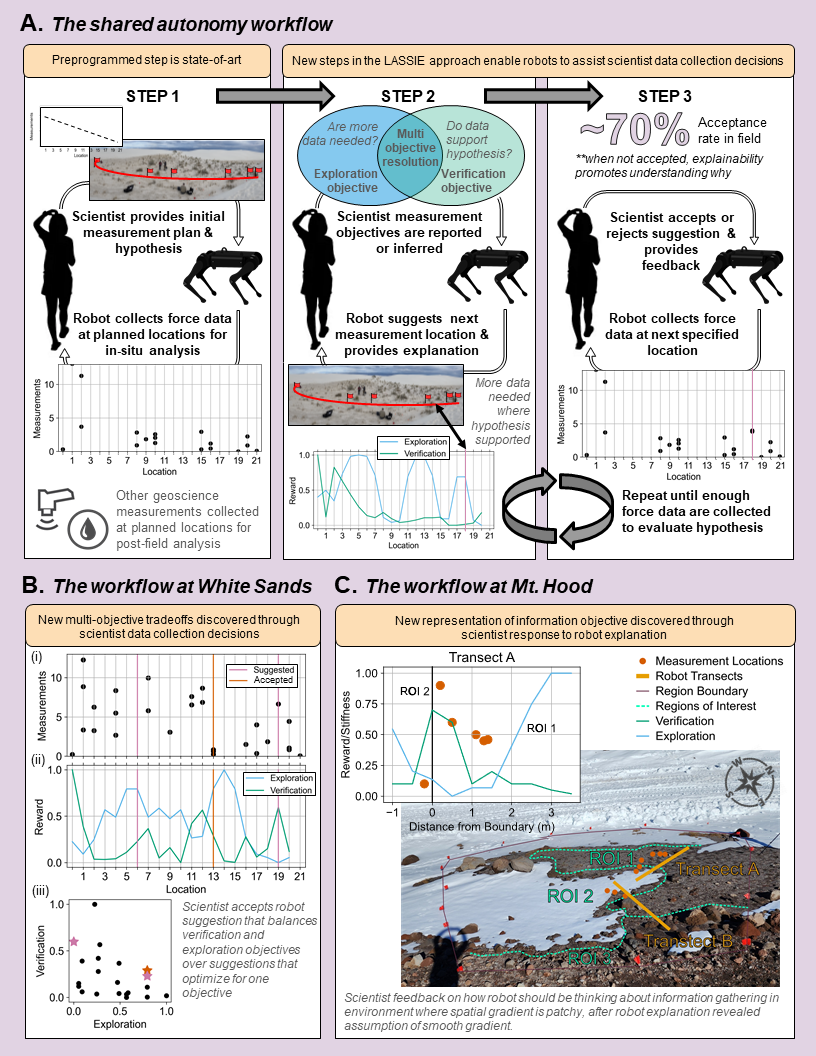} 
 \caption{\textbf{The human-robot shared autonomy workflow for scientific data collection decisions.}
 }
 \label{Fig.Ops}
 \end{figure}
 \addtocounter{figure}{-1} 
\begin{figure}[t!]
    \centering
    \caption{(Continued from previous page.) (\textbf{A}) Step 1: Scientist makes plan of where to initially collect measurements and robot executes plan to measure regolith mechanical responses by walking. Step 2: Scientist measurement objective priority is reported or inferred by robot based on a human model, which the robot uses to generate explainable suggestions for the scientist about where to take measurements next. Step 3: Scientist accepts or rejects the robot suggestions, and provides feedback to improve robot suggestions in the future. Robot collects data from new specified location, and Steps 2 and 3 are continuously cycled until the scientist is ready to stop data collection. (\textbf{B}) Results from White Sands illustrated in the (i) measurement, (ii) reward, and (iii) objective function spaces. In (i), black circular markers represent the existing measurements along the transect; the pink vertical lines represent robot suggested locations and the orange vertical line represents scientist-accepted location. In (ii), the blue curve represents the reward values corresponding to the exploration objective and the green curve represents the reward values corresponding to the verification objective. In (iii), black circles represent available measurement locations; pink and orange stars correspond to the robot-suggested locations and scientist-accepted location, respectively. (\textbf{C}) Results from Mt. Hood illustrates a scientist's sketch of the different environment gradient patches -- composition (darker, lighter/red), moisture (wet, dry), grainsize (fine grains, fine grains with cobbles, ultra fine grains, cobble and boulder) -- and their preferred data collection strategy for evaluating patchiness.}
\end{figure}

\begin{figure}
\centering %
\includegraphics[width=0.85\textwidth]{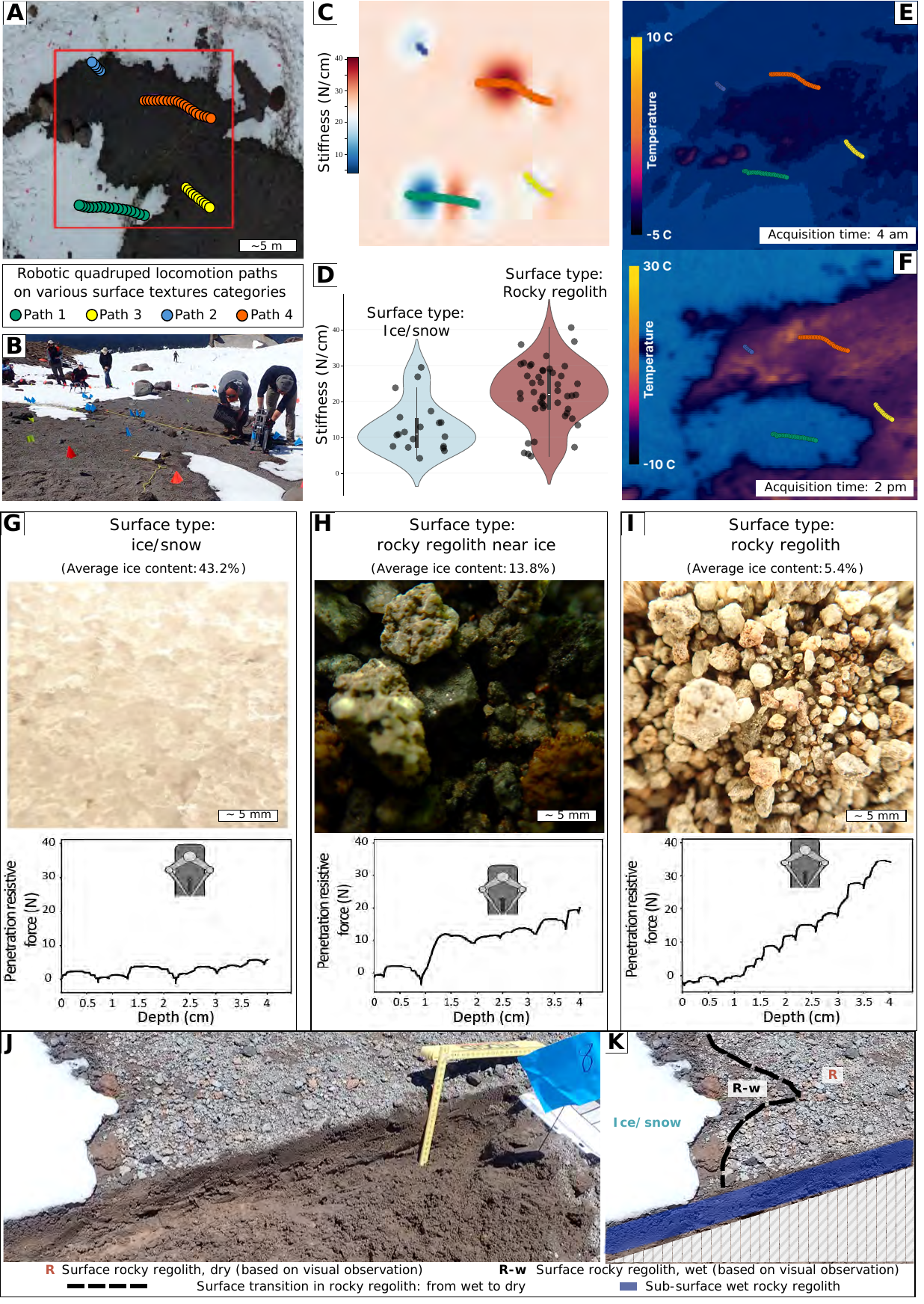} 
\caption{\textbf{Field measurements at Mt. Hood, OR.}
}
\label{Fig.MH}
\end{figure}
 \addtocounter{figure}{-1} 
\begin{figure}[t!]
    \centering
    \caption{(Continued from previous page.) (\textbf{A}) Top view drone image and (\textbf{B}) in-situ field view of the sampling region and robotic paths across icy and rocky terrains. (\textbf{C}) High-resolution map of local penetration resistance. (\textbf{D}) Distribution of resistive forces across terrain types. Drone-acquired thermal images in the (\textbf{E}) morning and (\textbf{F}) afternoon. (\textbf{G-I}) Representative force–depth profiles measured on three distinct surface types, shown with corresponding macro-scale surface images. (\textbf{J-K}) Trench exposing the shallow subsurface across an ice–regolith transition. }
\end{figure}

\begin{figure}
\centering %
\includegraphics[width=0.85\textwidth]{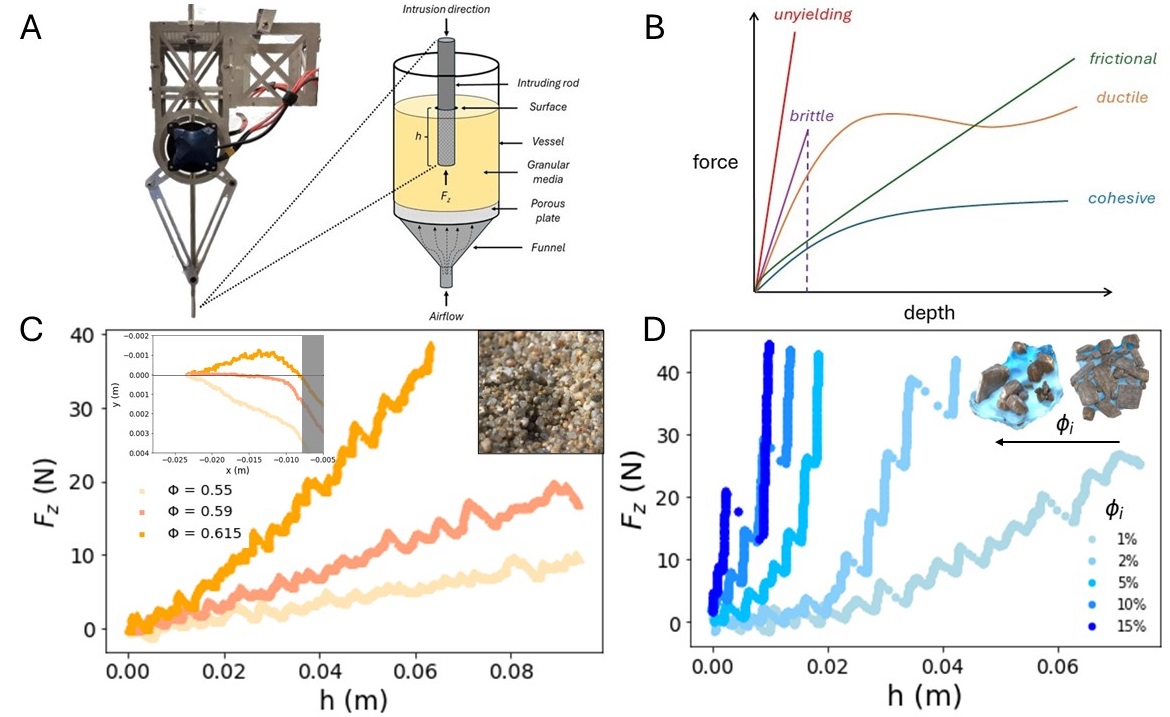} 
\caption{\textbf{Laboratory robotic intrusion tests into various granular materials and admixtures.} (\textbf{A}) Robotic intruder leg used in experiments, actuated with two direct-drive gearless motors, which is driven into the adjacent sketch of the experimental air-fluidizing apparatus. (\textbf{B}) Conceptual framework relating qualitative rheological trends to the range of observed behaviors encountered in various granular mixtures. (\textbf{C}) Force-depth profiles for sand prepared at $\phi_{min}$, $\phi_{c}$, and $\phi_{max}$, where each profile is an average of four intrusion tests. Inset: surface displacement profiles. The horizontal dotted line at y = 0 m represents the granular surface. The transparent grey bar symbolizes the intruder location. Negative y values reflect material that has dilated and extended above the surface, while positive y measures correspond to material that has been compacted. Values of x approaching 0 are closer to the center of intrusion. ((\textbf{D}) Representative force
profiles for samples prepared at $\phi_{i}$ = 1, 2, 5, 10, and 15$\%$.}
\label{RuckDougFig}
\end{figure}

\begin{figure}
\centering %
\includegraphics[width=1\textwidth]{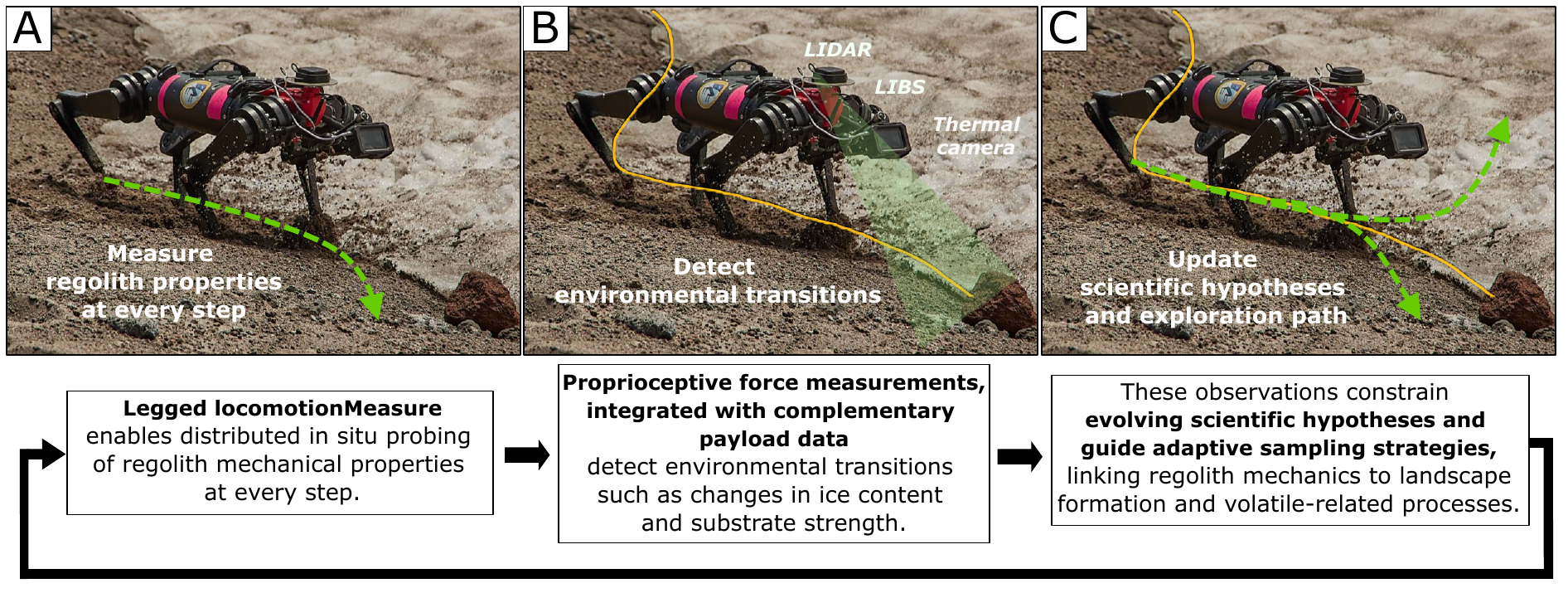} 
\caption{\textbf{Mechanically informed, hypothesis-driven surface exploration.} (\textbf{A}) Legged locomotion enables distributed in situ probing of regolith mechanical properties at every step. 
(\textbf{B}) Proprioceptive force measurements, integrated with complementary payload data (e.g., LiDAR and thermal imaging), detect environmental transitions such as changes in ice content and substrate strength. 
(\textbf{C}) These observations constrain evolving scientific hypotheses and guide adaptive sampling strategies, linking regolith mechanics to landscape formation and volatile-related processes.
}
\label{Fig.vision}
\end{figure}


\clearpage 

%
\bibliography{LASSIE_ScienceAdvancesSubmission} 
\bibliographystyle{sciencemag}

%
%
%
%
%
%


\section*{Acknowledgments}
Special thanks to the Timberline Lodge staff at Mt. Hood, especially Miles Bland, and the park staff at White Sands, especially David Bustos, for their support with permitting, access, and logistics during field operations. 
Thanks to student LASSIE team members Zachary Lee, Deanna Flynn, Sean Buchmeier, Liam Bouffard, Freya Whittaker, Mason Allen, Diana Garcia, Lucky Kovvuri, Iris Li, Natalie Jones, Ian Brunton, Tatiana Gibson, Sharissa Thompson, Jeongwoo Cho, and Eric Sigg for their assistance with field operations, and to Justin Durner and Sean Grasso for capturing photo and video of field operations.

\paragraph*{Funding:}
This work was funded by the NASA Planetary Science and Technology Through Analog Research (PSTAR) program, Grant 80NSSC22K1313, and NASA Mars Exploration Program (MEP) Technology Development Funds.

\paragraph*{Author contributions:}
CGW, MN, and FQ designed and led field operations, made the primary writing contributions, and led the generation of figures and video. SL, JGR, and JDC participated in field operations and made significant contributions to writing and the generation of figures and the video. BEM, YZ, JMB, JB, AME, EF, EBH, ICR, JJSC, and SS participated in field operations, led data analysis efforts, made significant contributions to the generation of figures, and contributions to writing. MRZ, RCE, KFR, DJJ, DEK, FRH, and TFS supported the design of field operations, participated in operations, and made contributions to writing and the generation of figures. 

\subsection*{Supplementary materials}
Materials and Methods\\
Fig. S1\\
References \textit{(84-\arabic{enumiv})}\\ 
Movie S1


\newpage


\renewcommand{\thefigure}{S\arabic{figure}}
\renewcommand{\thetable}{S\arabic{table}}
\renewcommand{\theequation}{S\arabic{equation}}
\renewcommand{\thepage}{S\arabic{page}}
\setcounter{figure}{0}
\setcounter{table}{0}
\setcounter{equation}{0}
\setcounter{page}{1} 


\begin{center}
\section*{Supplementary Materials for\\ \scititle}

Cristina G. Wilson$^{1\dagger}$, Marion Nachon$^{2\dagger}$, Shipeng Liu$^{3}$, John G. Ruck$^{4}$,  
\\ J. Diego Caporale$^{3}$, Benjamin E. McKeeby$^{5}$, Yifeng Zhang$^{3}$, Jordan M. Bretzfelder$^{6}$, 
\\ John Bush$^{3}$, Alivia M. Eng$^{6}$, Ethan Fulcher$^{3}$, Emmy B. Hughes$^{6}$, Ian C. Rankin$^{1}$, 
\\ Jelis J. Sostre Cortés$^{6}$, Sophie Silver$^{4}$, Michael R. Zanetti$^{7}$, Ryan C. Ewing$^{8}$,
\\ Kenton R. Fisher$^{8}$, Douglas J. Jerolmack$^{4}$, Daniel E. Koditschek$^{9}$,
\\ Frances Rivera-Hernández$^{6}$, Thomas F. Shipley$^{10}$, and Feifei Qian$^{3\ast}$
\\
\small$^\ast$Corresponding author. Email: feifeiqi@usc.edu\\
\small$^\dagger$These authors contributed equally to this work.

\end{center}

\subsubsection*{This PDF file includes:}
Materials and Methods\\
Figure S1\\
Captions for Movie S1\\

\subsubsection*{Other Supplementary Materials for this manuscript:}
Movie S1\\

\newpage


\subsection*{Materials and Methods}

\subsubsection*{Robotic Legs as Scientific Sensors to Measure Regolith terramechanical Properties in High Resolution}\label{sec:method-legs}

To create the next-generation legged rovers that can agilely move across a wide variety of challenging planetary surfaces while effectively collecting information about the environment, we developed a sensing framework that enables robotic legs to function as distributed geotechnical probes of regolith mechanics. 

As a first step towards this ``sampling during locomotion'' capability, we designed a standalone robotic leg (Fig. \ref{Fig.7}A) that functions as a field-deployable scientific instrument capable of sensitively measuring regolith resistive forces~\cite{bush2023robotic}.
The leg is a 2-degree-of-freedom (DoF) system actuated by two direct-drive (\ie gearless) brushless DC motors, precisely controlled by an ODrive v3.6 motor controller. Direct-drive (or quasi–direct-drive, with gear ratios $<$10:1) actuators minimize backlash and friction, providing high transparency and proprioceptive sensitivity~\cite{kenneally2018actuator}.
By integrating joint kinematics with motor current measurements, the leg continuously estimates external interaction forces at 100 Hz. 

We integrated this proprioceptive sensing with granular physics to extract regolith terramechanical properties from the leg force signals. 
In granular media, penetration resistance reflects packing fraction, particle-scale friction, and effective bearing capacity, and therefore provides a quantitative descriptor of substrate state. The robotic leg measures penetration resistance by recording normal contact force as a function of insertion depth during a controlled vertical penetration (which we refer to as the ``vertical penetration measurements'' in main text). The slope and shape of the resulting force–depth curve directly encode regolith strength and compaction state. 
Building on this physical interpretation, we implemented sensing-oriented regolith-probing gaits on the Traveler leg, together with algorithms to interpret and visualize regolith properties in situ.
Figure \ref{Fig.7}D illustrate the field user interface used to command the leg and visualize regolith properties in situ. The interface enables operators to set different gait trajectories, including ground detection, penetration, and shear. Operators can adjust key parameters for each gait—such as penetration angle, velocity, and duration—before execution. During motion, the interface provides real-time visualization of force and depth measurements, allowing users to assess local regolith properties as the robot moves.

Building upon the standalone leg, we further extended the regolith sensing capability to a quadruped robot, Spirit (Fig. \ref{Fig.7}B, C), to enable the robot to measure regolith properties at each step along a continuous path in natural environments. Unlike the standalone leg with a stationary base~\cite{qian2019rapid,bush2023robotic,ruck2024downslope}, Spirit’s body undergoes translation and rotation during locomotion, introducing dynamic accelerations that can bias force estimates. Accurate recovery of substrate reaction forces therefore requires simultaneous estimation of body pose and ground-plane orientation. To address this, we deployed two sensing-during-locomotion gaits developed in our recent work~\cite{fulcher2025effect}, together with a ground-plane–frame–based force estimation method that compensates for body motion during probing under both quasi-static and dynamic gaits.

The \crawlgait gait (Fig. \ref{Fig.7}B) operates in a quasi-static regime, minimizing inertial forces; each step consists of two phases--a penetration phase, where one leg serves as a vertical penetrometer and directly measures substrate reaction forces while the other three maintain continuous ground contact for stability, and a transition phase, where the robot shifts its center of mass forward before the next sensing leg begins the following cycle. The ground-plane frame is determined from the three-leg support configuration, and the proprioceptively estimated contact forces measured by the probing leg were transformed into this frame using the same ground estimation. Thus, the regolith properties can be characterized without relying on additional exteroceptive sensors.

In addition to the quasi-static crawl gait, we employ a fast \trotgait gait (Fig. \ref{Fig.7}C) to rapidly survey large terrain regions. Unlike the crawl gait, the trotting gait operates in a dynamic regime and accounts for inertia through momentum-based force estimation. Because only diagonal legs are in contact during trotting, ground-plane estimation relies on post reconstruction rather than direct geometric support. Although this approach yields less accurate regolith measurements, it enables rapid, continuous sensing and provides a complementary trade-off between spatial coverage and measurement fidelity. Together, these sensing modes establish a multi-scale measurement framework in which regolith mechanics can be characterized either with high precision (crawl) or high spatial coverage (trot).

Our approach offers several key advantages. First, by recognizing that every step is an opportunity to collect data, the robot opportunistically measures regolith strength at high spatial resolution without additional effort. Secondly, the legs can serve as flexible active sensors, capable of switching probing modes to extract complementary terramechanical information. These measurements reduce the risk of catastrophic locomotion failures by enabling real-time substrate assessment. Furthermore, because the method relies only on intrinsic motor sensing --- without additional wiring or fragile external sensors --- it increases robustness against damage from impacts or dynamic maneuvers, making it well-suited for long-term field deployment.

\subsubsection*{Cognitively-Inspired Algorithms to Support Human-Robot Shared Autonomy of Data Collection Decisions}\label{sec:method-decision}

To enable real-time adaptations of robotic locomotion and exploration plans in response to incoming measurements, we created cognitively-inspired algorithms that allow robots to participate in the decision making process with human scientists to interpret observations and dynamically adjust exploration plans.

To develop the algorithms, we ran a series of cognitive-behavioral studies on how expert Earth and planetary scientists connect high-level science objectives to low-level data collection plans, and how they update objectives and adapt data collection plans in response to incoming information. 
In the seminal work \cite{wilson2020}, we found scientists initially rely on simple, efficient heuristics endorsed by environmental sampling authorities to fulfill an exploration (\ie  information-gathering) objective: equal-interval sampling along a gradient of interest to avoid location bias \cite{epa2002,wolman1954method} and taking a uniform number of measurements at each location to capture natural variability and handle measurement error \cite{geboy2011quality, osti_6665284}.
Subsequent work \cite{liu2023understanding} showed scientists eventually transition (once sufficient information is gathered) from an exploration objective to a hypothesis verification objective, where the aim of data collection is to improve confidence in the hypothesis to a desired level by verifying areas of high or low hypothesis discrepancy.
From this, we developed algorithmic representations of the exploration and verification data collection objectives that accurately predicted over 90\% of expert scientist selected data collection Locations along a Transect, across all simulation-based and field studies with scientists.

We have embedded the new data collection algorithms within a shared autonomy workflow, where the robot takes on increased responsibility for data collection decisions. 
Initially, the workflow mimics state-of-the-art teaming: the human scientist selects preliminary Locations along the Transect to collect data, and the robot executes the force measurements at these Locations and transmits the data back to the scientist (Fig. \ref{Fig.Ops}A, Step 1).
The novel contributions of our workflow occur after this preliminary data collection, when the robot takes on the responsibility of generating candidate data collection Locations, and the human serves as a sampling authority, approving or rejecting candidates.

The robot generates candidate data collection Locations based on the scientist's current objective, exploration or verification, and running the corresponding algorithm (Fig. \ref{Fig.Ops}A, Step 2). 
The scientist's objective priority can be determined through self-report, or inferred by the robot based on a model of how scientists transition and weight objectives: \eg the model discovered in \cite{liu2023understanding} that scientists favor exploration in early sampling when information is low, and verification in later sampling steps.
Robot data collection Location suggestions are explainable based on the reported/inferred objectives: \eg the robot suggests sampling at location \textit{X} to optimize an exploration objective of increasing information coverage, the robot suggests sampling at location \textit{Y} to optimize a verification objective of testing areas of high hypothesis discrepancy.

Scientists' accept or reject robot candidate Locations and provide feedback on explanations that can improve the data collection algorithms and/or model (see Fig. \ref{Fig.Ops}A, Step 3). 
The learning occurs when there are discrepancies between the robots suggested Location and rationale and scientists decisions. Two types of discrepancies are important and lead to updating objective and strategy models. A scientist might agree with the explanation of data collection objective, but reject the suggested candidate Location because there is a different candidate Location that is better aligned with the objective: \eg yes, my objective is exploration, but Location \textit{X} is not the best for improving information coverage.
Alternatively, a scientist might accept the suggested candidate Location, but disagree with the explanation: \eg yes, I want to sample from Location \textit{Y}, but not because my objective is to test areas of high hypothesis discrepancy. 

The shared autonomy workflow allows robots to independently identify scientifically valuable data collection targets, and therefore allows for greater flexibility in data collection in response to new observations than workflows where scientists are solely responsible for decision making and robots serve as mobile sensor suites. 
The current version of the shared autonomy workflow has scientists supervising robot data collection decisions at every new Location, but with iterative testing and improvement from scientist feedback, we believe it will be possible to progressively reduce the need for human supervision.
Therefore, in the long term, the shared autonomy workflow will reduce the burden of low-level repetitive data collection decisions from humans, allowing them to focus on higher-level tasks that require their full expertise, ultimately enhance science outcomes.

\subsubsection*{Fieldwork Campaigns: Geoscience Instrumentation and Workflows}\label{sec:geo-instruments}

To characterize physico-chemical properties of the field areas' surface and shallow-subsurface, and to support the interpretation of robotic locomotion data, geoscience measurements were collected at several spatial scales (ranging from Regions, to Transects, to Locations). Regions range from kilometric to decametric in size, and were selected by geoscientists to be representative of the sites' diversity in landscape features and surface textures.
In each Region, geoscientists identified key Transects that captured a gradient of surface textures; Transects ranged from 5 to 150 meters long depending on the variability of surface textures observed.
Geoscientists specified a hypothesis about how robot leg measurements would change along a given Transect and they chose an initial set of Locations where to acquire measurements both with the legged-robot (which provided force data of the legged penetration into the substrate) and with geoscience instrumentation (which provided physico-chemical properties of the substrate) to evaluate the hypothesis. 

The force measurements taken via robot legs at the initial Locations were used in the shared autonomy workflow to generate robot suggestions for   Locations along the Transect to collect additional data (see Figure~\ref{Fig.Ops}A).
These robot suggestions to the geoscientists were produced through algorithms inspired by how expert scientists make decisions, balancing objectives of information gathering (exploration) and hypothesis evaluation (verification) \cite{wilson2020,liu2023understanding,liu2024modelling}.
The geoscientists could accept the robot suggestion, or reject it and choose a different Location. 
In either case, the battery of measurements was taken at the next Location and, subsequently, a new set of robot-suggested Locations were generated. 
This process continued until the geoscientists felt they had enough data to make a conclusion about the hypothesis at the Transect, at which point we moved to the next Transect or Region.

Geoscience measurements were acquired at Region-, Transect-, and Location-scales to characterize physico-chemical properties of the surface and shallow (centimeter-deep) subsurface, and to enable comparison with the force data of the robotic-legged rheology measurements (Section ~\ref{sec:geo-instruments}, Fig. \ref{Fig.7}).
The suite of geoscience measurements acquired varied across the analogue field sites, based on the particular Regions' characteristics, and included a selection of: topographic data (via ground LiDAR), compositional analyses (via spectroscopic techniques), drone images, thermal infrared data, context and high-resolution images of the geologic textures, and soil moisture content. Moreover, centimetric-deep mini-trenches were dug at specific Locations to expose and image the shallow subsurface, and samples were collected for \textit{a posteriori} laboratory-based analyses (e.g. grain-size assessment). 
Trenching and sampling were performed last, to minimize disturbance to surface and near-surface measurements. 

In the following sections, we first summarize the geoscience instrumentation employed, and then describe the detailed methodological workflows implemented during the White Sands and Mt.~Hood field campaigns.

\paragraph{Geoscience Instrumentation.}
The geoscience instrumentation suite used during fieldwork included analytical techniques that are standard in geosciences for characterizing textural and compositional properties of the substrate, and also included analytical techniques analogous to instruments flown on space missions, in particular onboard the two most recent NASA Mars rovers, \textit{Curiosity} and \textit{Perseverance}.

\subparagraph{Drones Images in the Visible and Thermal-Infrared Spectral Ranges.}
At our Mt. Hood field site, we deployed drones equipped with visible and thermal infrared (TIR) cameras to capture high-resolution image data of the surface (Fig.\ref{Fig.overview}, \ref{Fig.7}). The TIR data were collected using a FLIR Vue R camera custom-mounted on a DJI Phantom platform, providing a ground sampling distance of approximately 20 cm per pixel. Flights were conducted every three hours over a three-day period to capture the full diurnal temperature curve of the surface. This broadband thermal radiometry enables mapping of apparent surface temperature variations, which can be used to infer thermal inertia and related properties such as ice content, particle size, and degree of cementation (e.g. \cite{Fergason2006, bandfield2008, koeppel2024}). Such methods parallel orbital thermal observations from instruments like THEMIS on Mars Odyssey \cite{christensen2004} and Diviner on the Lunar Reconnaissance Orbiter \cite{paige2010}, which have been used to study surface composition and thermophysical variability on Mars and the Moon. The deployment of the drones was restricted to the field sites where regulations authorized their use.

\subparagraph{Topography.}
We used LiDAR (Light Detection and Ranging) to capture surface topography and provide contextual 3D models of the terrains (Fig.\ref{Fig.overview}, \ref{Fig.7}). It is a technique employed onboard several spacecrafts missions for mapping the topography of planetary bodies, including the Earth, our Moon, asteroids, Mars, and Mercury (e.g., MOLA instrument onboard the Mars Global Surveyor mission; MLA instrument onboard the MESSENGER mission; BELA instrument onboard the BepiColombo mission \cite{smith2001, cavanaugh2007, thomas2021}).
We conducted LiDAR measurements via a ground-based instrument (Fig.\ref{Fig.7}), scanning selected terrains along our Transects, to acquire dense LiDAR point cloud data and generating Digital Elevation Models (DEMs) of the studied areas. We also tested the deployment of a backpack-carried LIDAR: the Kinematic Navigation and Cartography Knapsack (KNaCK), developed at Marshall Space Flight Center \cite{zanetti2023} and used to generate 4 cm/px DEMs (Fig.\ref{Fig.7}H).

\subparagraph{Macro-Images.}
To visually characterize the surface textures at the locations where the legged measurements were collected, we acquired high-resolution close-up photographs, using a compact digital camera equipped with a macroscopic mode.
The resulting images provide high enough resolution to allow visual inspection of the sand grains (including their sizes and arrangements) and were used to describe and classify the observed textures (Fig.\ref{Fig.WS}, \ref{Fig.MH}).

\subparagraph{Compositional Analyses via Handheld Spectrometers.}
We used several handheld instruments (Fig.\ref{Fig.7}), that allow rapid, in-situ acquisition of localized compositional data (and eliminate the need for sample collection for subsequent laboratory analyses) at Locations where legged measurements were performed. The analytical techniques we used are complementary, and each are sensitive to specific compositional properties, including elemental composition, and mineralogical characteristics. Moreover, these techniques are analogous to those used onboard several recent planetary missions.

Visible and Infrared (VisIR) spectroscopy was conducted using a SpectralEvolution PSR-3500 spectrometer to acquire passive reflectance measurements of surface materials (Fig.\ref{Fig.7}). This technique measures the intensity of reflected sunlight across the visible to shortwave infrared range (approximately 0.35-2.5 $\mu$m), allowing for non-destructive characterization of mineralogical and hydration features directly in the field. Reflectance spectroscopy is widely used in planetary and terrestrial studies because it captures diagnostic absorption features associated with specific minerals, including Fe-bearing silicates, phyllosilicates, and carbonates, as well as $H_2O$- and $OH$-related bands near 1.4, 1.9, and 2.2 $\mu$m (e.g. \cite{clark1990, mustard1998, bishop2008}). The passive nature of this approach enables repeated, in situ measurements without altering or disturbing the surface. Similar techniques have been fundamental to remote compositional analysis on planetary missions, both from spacecrafts (e.g. \cite{brown2004, bibring2017, kitazato2019}) as well as from onboard rovers including the Pancam instrument on the Mars Exploration Rovers \cite{bell2004}, Mastcam on the \textit{Curiosity} rover \cite{bell2017}, Mastcam-Z \cite{bell2021} and also the near-infrared spectrometer of the SuperCam instrument \cite{maurice2021, wiens2021, royer2023} on the \textit{Perseverance} rover, which operate in the visible to near-infrared range to identify minerals and assess hydration state on the Martian surface. 

The LIBS (Laser-Induced Breakdown Spectroscopy) technique enables the determination of a material's elemental composition, over a sub-mm analysis spot, using an active spectroscopy method: a focused laser beam ionizes the target material, generating a plasma whose emitted light is collected by spectrometers \cite{cr2013}. Each chemical element present in the ionized material produces characteristic spectral emission lines, allowing for their identification.
The LIBS technique was first used on a space mission in 2012 as part of the ChemCam instrument onboard NASA's \textit{Curiosity} Mars rover \cite{wiens2012, maurice2012}. It was later selected again for the next generation of NASA Mars rovers, with the SuperCam instrument onboard the \textit{Perseverance} rover \cite{wiens2021, maurice2021}.

\subparagraph{Moisture / Water Content.}
At the White Sands field site, we assessed variations in moisture content within the sandy substrate by collecting resistivity measurements using a Sinometer VC97 Digital Multimeter probe. The probe was inserted into the shallow sub-surface at locations and depths similar to those of the robotic locomotion measurements. Because resistivity serves a a proxy for moisture content, these measurements provide insights into the relative capability of the geologic material to retain humidity. 
At the Mount Hood field site, we evaluated water content of the icy volcanic regolith through a sample collection protocol. Material was collected then heater to melt and evaporate the interstitial ice and water, and the samples were weighed before and after boiling to quantify overall water content.

\subparagraph{Sample Collection for Laboratory Analyses.}
Samples were collected, and grain-size analyses performed on samples collected and run via a Beckman-Coulter LS-13-320 Laser Diffraction Particle Size Analyzer.

\subparagraph{Mini-Trenches for Inspection of Shallow Subsurface.}
To visually examine the shallow sub-surface with which the robotic legs interacted during intrusion, we excavated small trenches, typically 5 to 10 cm deep, at several locations after completing our series of measurements.
At the White Sands field site, these mini-trenches enabled us to visually identify the presence, location, and distribution of stratification and textural variations. At the Mt Hood site, they enabled visual identification of ice within the subsurface. This information was useful for correlating subsurface conditions with the force measurements and for interpreting variations in the measured forces.

\paragraph{Workflow at the White Sands National Park Field Site.}
Our fieldwork at the White Sands site focused on several distinct Regions of interest, that we investigated following a similar methodology. The investigations conducted in Region 1 (Fig.\ref{Fig.fieldwork}), were carried out following the workflow below.

Region 1 was selected based on the geoscience hypothesis that interdune areas between barchan dunes host smooth spatial gradients in surface textures, which robotic legs could detect during locomotion. This hypothesis was initially developed through the examination of satellite images of the White Sands National Park dune field, which were used to identify representative dune types and to assess the overall morphology and scale of the interdune areas. It is also based on general observations of dune fields on Earth and other planetary bodies, particularly Mars, where satellite images typically provide the first datasets used to characterize surface units and guide planetary exploration.

During fieldwork, the first step in investigating Region 1 involved geoscientists visually identifying the main surface textures present, along the expected gradient between two dunes. These naked-eye observations revealed a diversity of surface textures ranging from loose to encrusted materials, which were classified into several categories (Fig. \ref{Fig.WS}A).

Based on this initial reconnaissance, geoscientists selected a Transect (55 meters long) extending from the lee face of one dune, across the interdune area, to the windward stoss face of the next dune. Along this Transect, they selected 20 initial Locations that represented the main categories of observed surface textures observed as well as locations where these textures transitioned.
In terms of decision-making process, although the number of Locations was not strictly constrained, the team considered the time required for data acquisition and selected a number that could be realistically completed within a single fieldwork day.

Following the selection of the Transect and the initial Locations, we collected the suite of robotic  and geoscience measurements, prioritizing co-location, and following a sequence of activities that minimized surface disturbance prior to data collection. 
At each of the 20 initial Locations (marked with flags for visual identification) we acquired non-destructive geoscience measurements, followed by robotic legged data. 
These robotic leg measurements were subsequently used in the shared autonomy workflow to generate robot suggestions for Locations along the Transect to collect additional data.
Following the investigations at the individual Locations, the quadruped legged robot was deployed, to acquire Force measurements during locomotion using a \trotgait gait.

\paragraph{Workflow at the Mount Hood field site.}
The fieldwork investigations conducted at the Mount Hood site (Fig.  \ref{Fig.fieldwork}) were carried out following the workflow below.
First, human operators conducted a visual inspection of the site and identified Regions where ice-covered surfaces transitions into patches of exposed rocky regolith (Fig.~\ref{Fig.MH}). A technical constraint of our field workflow at this site was that the selection of Regions of interest could not be pre-planned far in advance due to daily variations in ice coverage driven by melting.
For each Regions of interest, human operators established a region boundary (by planting flags), delineated the boundaries between the main categories of surface textures (ice-covered surface; rocky regolith, of varying sizes and tones), and selected a set of Transects or paths (spanning metric to decametric scales) that either crossed visible surface-textures transitions or explore the lateral variability within a single surface-texture (e.g. rocky regolith without ice cover, along a path parallel to the local iced-covered zone).

The suite of robotic and geoscience measurements was then collected in a manner similar to that used at the White Sands site: the robotic measurement was performed by the quadruped using a \crawlgait gait, and coupled with a suite of geoscience measurements, to characterize the surface and shallow subsurface, both at the Regional and at Locations scales. 
Notable differences in instrumentation deployed, compared to the fieldwork campaign at White Sands, included the use of drone (for images acquisition at different wavelength), and the evaluation of overall water content in the rocky regolith (using a jet boil). With authorization to operate drones at this site, we acquired images in both the visible and the thermal infrared ranges at different times, to provide contextual views of the Regions and track temporal changes in ice coverage due to melting (Figs. \ref{Fig.MH}, \ref{Fig.7}).
The water content evaluation constituted a ground-based way to quantitatively assess the amount of ice present at specific Locations, and that might affect the robotic Force measurements (\ref{Fig.MH}).

\subsubsection*{Lab Experimental Setup to Connect Regolith Force Measurements from Analog Missions}\label{sec:LegSense-lab}

Granular materials exhibit complex mechanical behavior, transitioning between fluid-like and solid-like states depending on packing fraction, grain properties, friction, and cohesion \cite{behringer_chakraborty}. Variations in volume fraction $\phi$, grain size and shape, interparticle friction, and cohesive forces produce measurable changes in material strength and stability. Because vertical intrusion provides a controlled and interpretable perturbation, intrusion dynamics have become a practical method for characterizing granular rheology via macroscopic force laws. Here, we employ the standalone robotic leg as a laboratory rheometer to quantify how regolith mechanical response depends on controlled material parameters, particularly packing fraction $\phi$ and ice content. This approach establishes a mechanistic bridge between laboratory-controlled materials and force--depth signatures observed during field deployments.

\paragraph{Cohesionless Granular Preparation.}
Experiments were conducted in a custom air-fluidized chamber (diameter 21.6~cm) to enable reproducible preparation of granular packs at controlled volume fractions (Fig.~\ref{RuckDougFig}A) \cite{ruck2024downslope}. Fluidization resets the bed between trials and allows tuning of $\phi$: $\phi_{min}$ (random loose packing) was achieved by fully fluidizing then abruptly stopping airflow; $\phi_{c}$ by gradual pulsed airflow to compact the bed; and $\phi_{max}$ (random close packing) via mechanical tapping. Airflow and vibration were halted before height measurement and intrusion to ensure static initial conditions.

\paragraph{Ice--Granular Mixture Preparation.}
To isolate the mechanical contribution of ice in granular--ice mixtures, we prepared frozen samples with controlled ice fraction $\phi_i$. We adopted an ice-cemented preparation method commonly used in permafrost and icy soil geotechnical studies \cite{gertsch2006,gertsch2008,atkinson_zacny,atkinson,liu,mantovani}, consistent with morphologies observed at the Mount Hood field site. Natural quartz sand (200--300~$\mu$m) was cooled to $-20^\circ$C. A prescribed volume of water was mixed into 4{,}770~cm$^3$ of sand and homogenized through repeated tumbling to prevent clumping. The mixture was sieved into an acrylic chamber, vibrated to level and relax the surface, yielding a bulk volume fraction of $0.59$. Samples were frozen for 24~hours prior to testing. Intrusion tests were conducted immediately after removal from the freezer, with five penetration tests per sample at 2~cm/s to a depth of 10~cm. Ice fraction $\phi_i$ was systematically varied from 0--15\%. Thermistors monitored sample temperature during testing. All robotic hardware and sensing protocols matched those used in field experiments \cite{ruck2024downslope}. This unified preparation and sensing protocol allows direct comparison of force--depth responses across dry, cohesive, and ice-cemented regimes using a single instrumentation framework.

\paragraph{Quasi-static Vertical Intrusion Protocol.}
All intrusion experiments were conducted at a constant speed of $v = 2$~cm/s, well within the quasistatic regime (below the characteristic grain settling velocity $v_c = \sqrt{2gd} \sim 10$~cm/s, where $d$ is mean grain diameter) \cite{roth2021constant}. This ensures inertial effects are negligible and that the force response reflects material strength rather than dynamic drag. The intruder tip was an aluminum cylinder (diameter 2.54~cm for cohesionless tests; 0.71~cm for frozen-mixture tests), actuated by two high-torque brushless DC motors with position control (ODrive v3.6). Each test consisted of vertical penetration to a maximum depth $h_{max} = 10$~cm, followed by controlled withdrawal. Force measurements were acquired at 300~Hz using motor current and joint kinematics to estimate vertical reaction forces throughout intrusion \cite{kenneally,kenneally_2}, yielding continuous, high-fidelity force--depth curves for rheological interpretation.


\begin{figure}[tbhp]
\centering %
\includegraphics[width=0.85\textwidth]{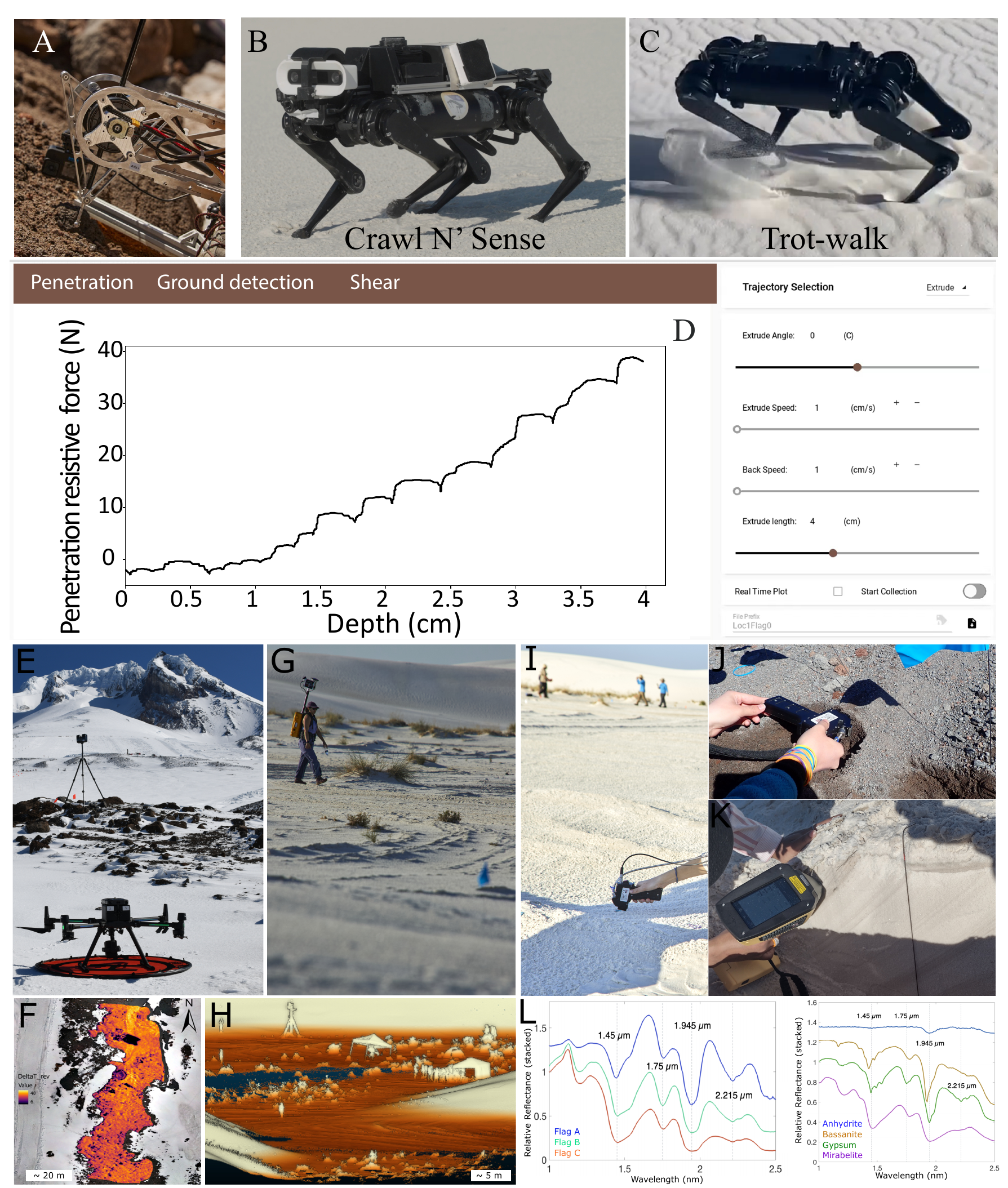} 
\caption{\textbf{Robotic platforms and scientific instruments deployed at field campaigns.} }
\label{Fig.7}
\end{figure}
 \addtocounter{figure}{-1} 
\begin{figure}[t!]
    \centering
    \caption{(Continued from previous page.) (\textbf{A}) Traveler, robotic leg rheometer for measuring regolith mechanics. Two quadrupedal ``sensing during locomotion'' gaits: (\textbf{B}) \crawlgait and (\textbf{C}) \trotgait. (\textbf{D}) Real-time control and data-collection interface showing force, speed, and position measurements during leg--terrain interaction experiments. (\textbf{E}) Drone equipped with cameras, and ground-based LiDAR.(\textbf{F}) Drone images: thermal infrared (TIR) data superposed on visible image. (\textbf{G}) LiDAR backpack-carried, KNaCK, and corresponding (\textbf{H}) topographic data model of the field area. Handheld spectrometers used to acquire (\textbf{I}) compositional data of the surface and of the (\textbf{J-K})  shallow subsurface in mini trenches. (\textbf{L}) ViSIR spectral data of three flag locations (left), and of four mineral standards (right) for comparison.}
\end{figure}


\clearpage 

\paragraph{Caption for Movie S1.}
\textbf{Legged Autonomous Surface Science in Analogue Environments.}
This video demonstrates the LASSIE framework, in which legged robots act as in situ geotechnical probes during locomotion. Through field deployments at White Sands National Park, NM, and Mount Hood, OR, the robot measures regolith mechanical properties at each step via proprioceptive force sensing, enabling spatially dense characterization of substrate strength and structure. These measurements are integrated with co-located geoscience observations to interpret environmental processes such as dune stabilization and ice–regolith interactions. The video further illustrates a human–robot shared autonomy workflow, in which robot-generated suggestions guide adaptive data collection to test scientific hypotheses. Together, these capabilities demonstrate how using robotic legs as scientific instrumentation transforms locomotion into measurement, enhancing scientific discovery and operational decision-making in complex planetary analogue environments.



\end{document}